\documentclass[11pt,a4paper]{article}

\usepackage{a4wide}
\usepackage[utf8]{inputenc}
\usepackage[english]{babel}
\usepackage{authblk}

\usepackage[dvipsnames,table]{xcolor}
\usepackage[utf8]{inputenc}
\usepackage[english]{babel}
\usepackage{amsmath}
\usepackage{amsfonts}
\usepackage{amssymb}
\usepackage[linesnumbered,vlined,ruled,boxed,commentsnumbered,resetcount]{algorithm2e}
\usepackage{adjustbox}
\usepackage{comment}
\usepackage{csquotes}
\usepackage{float}
\usepackage{graphicx,nicefrac}
\usepackage{longtable}
\usepackage{lscape}
\usepackage{multirow}
\usepackage{paralist}
\usepackage{pgf}
\usepackage{rotating}
\usepackage{setspace}
\usepackage{subcaption}
\usepackage{tabu}
\usepackage{tikz}
\usepackage[a]{esvect}
\usepackage{hyperref}
\usepackage{xspace}
\usepackage{booktabs}
\SetCommentSty{itshape} 
\SetKwComment{tcp}{$\triangleright$ }{}
\SetKwInOut{Input}{input}
\SetKwInOut{Output}{output}

\newcommand{\ED}{\textsc{ed}}
\newcommand{\SED}{\textsc{sed}}

\newcommand{\Kpp}{$k$-means$++$\xspace}
\newcommand{\Km}{$k$-means\xspace}
\newcommand{\Kpar}{$k$-means$||$\xspace}
\newcommand{\norm}[1]{\left\lVert#1\right\rVert_2}
\newcommand{\remove}[1]{}

\usepackage{amssymb}

\begin{document}



\title{Accelerating the k-means++ Algorithm by Using \\ Geometric Information}


\author[1,2]{Guillem Rodríguez Corominas\footnote{\texttt{guillem.rodriguez.corominas@upc.edu}}}
\author[1]{Maria J. Blesa\footnote{\texttt{maria.j.blesa@upc.edu}}}
\author[2]{Christian Blum\footnote{\texttt{christian.blum@iiia.csic.es}}}
\affil[1]{Universitat Politècnica de Catalunya (UPC), Barcelona, Catalonia}
\affil[2]{Artificial Intelligence Research Institute (IIIA-CSIC), Bellaterra, Spain}


\maketitle

\begin{abstract}
In this paper, we propose an acceleration of the exact \Kpp algorithm using geometric information, specifically the Triangle Inequality and additional norm filters, along with a two-step sampling procedure. Our experiments demonstrate that the accelerated version outperforms the standard \Kpp version in terms of the number of visited points and distance calculations, achieving greater speedup as the number of clusters increases. The version utilizing the Triangle Inequality is particularly effective for low-dimensional data, while the additional norm-based filter enhances performance in high-dimensional instances with greater norm variance among points. Additional experiments show the behavior of our algorithms when executed concurrently across multiple jobs and examine how memory performance impacts practical speedup.
\end{abstract}


\section{Introduction}

The \Km clustering is a widely used method in data clustering and unsupervised machine learning, aiming to divide a given dataset into $k$ distinct, non-overlapping clusters. This division seeks to minimize the within-cluster variance.

The \Km clustering problem becomes NP-hard when extended beyond a single dimension~\cite{Aloise2009}. Despite this complexity, there are algorithms designed to find sufficiently good solutions within a reasonable amount of time. Among these, Lloyd’s algorithm, also referred to as the standard algorithm or batch \Km, is the most renowned~\cite{Lloyd1982}.

The \Km algorithm is one of the most popular algorithms in data mining~\cite{Wu2007, Ikotun2023}, mainly due to its simplicity, scalability, and guaranteed termination. However, its performance is highly sensible to the initial placement of the centers~\cite{Arthur2007}. In fact, there is no general approximation expectation for Lloyd’s algorithm that applies to all scenarios, i.e., an arbitrary initialization may lead to an arbitrarily bad clustering. Therefore, it is crucial to employ effective initialization methods~\cite{Franti2019}.

The \Kpp \cite{Arthur2007} algorithm is one of the most popular initialization methods, which has thus been integrated into many standard machine learning libraries. The algorithm starts with an arbitrarily chosen initial center among the input points. Then, subsequent centers are selected via $D^2$ sampling, i.e., using a weighted probability selection, where the probability of selecting a point is proportional to its squared distance from the nearest existing center. Thus, points farther to the current centers are more likely to be selected.

Beyond its role in initializing the \Km algorithm, the \Kpp algorithm has been used independently in applications across a variety of fields such as crime domain \cite{Aubaidan2014}, defect prediction \cite{Ozturk2015}, user value identification \cite{Wu2021}, geographical clustering \cite{Lee2021} and lithology identification \cite{Ren2022}, to name a few. Moreover, it has also been used in coreset construction \cite{Bachem2015}.

As proven in \cite{Arthur2007}, the algorithm is known to provide an approximation guarantee. More specifically, the \Kpp is never worse than $\mathcal{O}(\log k)$ competitive, which improves to $\mathcal{O}(1)$  (constant approximation) on very well-formed data sets \cite{Jaiswal2012}.

The computational complexity of the \Kpp algorithm is $\mathcal{O}(nkd)$, where  $n$ represents the total number of points,  $k$ is the number of clusters, and  $d$ is the dimensionality of the data. This complexity arises from the fact that, at each iteration, the full dataset must be traversed to determine whether the newly selected center is the closest to each point. Furthermore, the $D^2$ sampling similarly involves iterating over the entire dataset in the worst case. Although the algorithm has a linear scalability concerning each parameter, it can become impractical for large datasets. Many acceleration techniques have been proposed to address this issue.

In this paper, we introduce a novel acceleration technique using the Triangle Inequality along with supplementary geometric information, while adapting the sampling procedure.

\section{Related Work}

As previously stated, the computational cost of the \Kpp algorithm can become impractical for very large instances. Thus, many methods have been proposed to accelerate the algorithm.

The \Kpp algorithm goes through two phases at each iteration. First, when a new center is added---to determine whether or not the new one is the closest one---the distances from all points to the newly added center have to be recalculated. The computational cost of this phase comes from calculating the distance between them, which can be huge in high-dimensional spaces. Second, the $D^2$ sampling phase requires scanning the whole data, and due to its sequential nature, it becomes difficult to parallelize. Thus, the methods usually split in two: \emph{approximate methods}, which modify some of the aspects of the algorithm to be more easily computable while trying to obtain provably similar results; and \emph{exact accelerations}, which maintain the same base idea but try to optimize some of its aspects, thus yielding the same results as the standard approach.

In \cite{Bahmani2012}, a parallel implementation of the \Kpp algorithm was proposed, named \Kpar, which obtains the same guarantee in expectation as the standard approach. The main idea is to sample more points than necessary. The algorithm first oversamples the number of centers to $k \log n$, by performing $\mathcal{O}(\log n)$ rounds of sampling in which $k$ points are sampled in parallel. Then, the $\mathcal{O}(\log n)$ sampled points are clustered again using the standard \Kpp algorithm to select the final $k$ centers. Although the total complexity of the algorithm rises to $\mathcal{O}(ndk \log n)$, the speedup comes from the effective distribution of the work in a parallel process. Some authors have proposed enhancements to this method. For instance, in \cite{Hmlinen2020}, two new methods are proposed based on a divide-and-conquer type of \Kpar approach and random projections. In \cite{Raff2021}, both the Triangle Inequality and a dynamic priority queue are used to accelerate both the \Kpp and \Kpar algorithms. In \cite{Xu2014}, another approximation using MapReduce \cite{Dean2008} was proposed. This approximation uses a pruning strategy based on the Triangle Inequality to reduce the redundant distance computations. 

As mentioned, some methods have been proposed to replace the costly $D^2$ sampling with other more efficient sampling techniques expected to yield similar results, e.g., using coreset trees~\cite{Ackermann2012} or Markov Chain Monte Carlo sampling \cite{Bachem2016-sublinear}, which makes the algorithm run in sublinear time with respect to the number of data points. However, this approach requires assumptions on the data distribution. In \cite{Bachem2016-seeding}, the same authors proposed an improvement that produces provably good clusterings even without such assumptions. Additionally, in \cite{Cohen2020}, the distances between points are approximated using a multi-tree embedding followed by rejection sampling to approximate the original sampling. 

In \cite{Lattanzi2019}, the authors developed a variant of the \Kpp algorithm that achieves a constant approximation guarantee using a local search procedure. In \cite{Liang2022}, the running time for constant approximation is improved by using a distance oracle to approximate the distances. In \cite{Chan2021}, the authors proposed using random projections to reduce the dimensionality of the data, thereby accelerating distance computations by performing them in a much lower-dimensional space. More recently, in \cite{Zhang2023}, an acceleration of the exact \Kpp algorithm is proposed using the Lower Bound Based Framework to reduce the number of distance calculations.

\section{Preliminaries}

This section introduces concepts and methods to aid us in subsequent sections. The main intention behind these concepts is to reduce the computational time employed by the algorithm by avoiding unnecessary or more complex calculations.

\subsection{Distances and Metrics}

The Euclidean Distance (ED) represents the shortest path between two points in the Euclidean space. More specifically, given two points $\vv{x}$ and $\vv{y}$ in a $d$-dimensional Euclidean space $\mathbb{R}^d$, the ED is calculated as follows:
\begin{equation}
    \ED(\vv{x},\vv{y}) = \norm{\vv{x} - \vv{y}} = \sqrt{\sum_{j=1}^{d} (\vv{x}_j - \vv{y}_{j})^2}
\end{equation}

where $ \norm{\vv{x} - \vv{y}}$  denotes the $l^{2}$ norm of the vector difference between $\vv{x}$ and $\vv{y}$. As ED is a metric, it satisfies the Triangle Inequality (TIE), which states that the direct path between two points is the shortest. More specifically,
    \begin{equation}
        d(\vv{x},\vv{y}) \leq d(\vv{x}, \vv{z}) + d(\vv{z}, \vv{y})
    \end{equation}
As its name indicates, the Squared Euclidean Distance (SED) modifies the ED by squaring its value. This modification, while seemingly straightforward, significantly impacts various computational and analytical domains. The formal mathematical representation of the SED is as follows:
\begin{equation}
    \SED(\vv{x},\vv{y}) = \norm{\vv{x} - \vv{y}}^{2} = \sum_{j=1}^{d} (\vv{x}_j - \vv{y}_{j})^2
\end{equation}
Compared to the ED, the SED accentuates differences by squaring the distance. This holds particularly true for points that are far apart. A primary benefit of the SED over the ED is its computational efficiency. The SED simplifies calculations by omitting the final square root operation required in the ED, especially in high-dimensional spaces. Furthermore, it mitigates numerical precision issues with square root computations. However, the SED is \emph{not} a metric, as it does \emph{not} satisfy the Triangle Inequality principle.\footnote{For instance, let us assume the 2-dimensional space with vectors $\vv{x} = (0,0)$, $\vv{y} = (2,2)$ and $\vv{z} = (1,1)$. Then: $8 = 2^2 + 2^2 = d(\vv{x},\vv{y}) \nleq d(\vv{x},\vv{z}) + d(\vv{z},\vv{y}) = (1^2 + 1^2) + (1^2+1^2) = 4$}.
    
Despite not being a metric, the SED preserves the ranking of distances, as, given the non-negativity property of the ED, it holds that $\ED(\vv{x},\vv{y}) < \ED(\vv{x},\vv{z}) \iff \SED(\vv{x},\vv{y}) < \SED(\vv{x},\vv{z})$,
Thus, the SED can be used for \emph{ranking} points relative to a reference point akin to the ED. This characteristic makes the SED particularly suited for optimization problems, as minimizing the SED is equivalent to minimizing the ED, but with simpler computational requirements~\cite{Kaplan2011}.

\subsection{The Triangle Inequality}
\label{section:TIE}

The Triangle Inequality (TIE) is a fundamental property extensively utilized in enhancing the efficiency of clustering algorithms \cite{Hamerly2010} as well as in related fields such as speeding up the closest codework search process in Vector Quantization (VQ) \cite{Huang1990}.

In algorithms like the \Km, assigning a point to its nearest center is a major time-consuming task. For a given set of centers $\mathcal{C}$ and a point $\vv{p}$ in a $d$-dimensional space, the objective is to assign $\vv{p}$ to the closest center $\vv{c}_{\mathrm{best}} \in \mathcal{C}$, i.e., ensuring that $d_{\mathrm{min}} = d(\vv{p},\vv{c}_{\mathrm{best}}) \leq d(\vv{p},\vv{c})$ for all $\vv{c} \in \mathcal{C} \setminus \{\vv{c}_{\mathrm{best}}\}$., where $d_{\mathrm{min}}$ is the distance from point $\vv{p}$ to its closest center.

When $d$ is a metric, as per the TIE, the following holds true: $d(\vv{c},\vv{c}_{\mathrm{best}}) \leq d(\vv{p},\vv{c}) + d(\vv{p},\vv{c}_{\mathrm{best}})$. Given that $d(\vv{p},\vv{c}_{\mathrm{best}}) \leq d(\vv{p},\vv{c})$ by definition, we can substitute into the previous equation and obtain
\begin{equation*}
    d(\vv{c},\vv{c}_{\mathrm{best}}) \leq 2 \cdot d(\vv{p},\vv{c})  \enspace.
\end{equation*}

Therefore, any enter $\vv{c}$ can be disregarded if
\begin{equation} \label{eq:TIEFilter}
    d(\vv{c},\vv{c}_{\mathrm{best}}) > 2\cdot d(\vv{p},\vv{c}) \geq  2 \cdot d(\vv{p},\vv{c}_{\mathrm{best}}) = 2 \cdot d_{\mathrm{min}} \enspace,
\end{equation}
i.e., any center $\vv{c}$ meeting the condition $d(\vv{c},\vv{c}_{\mathrm{best}}) > 2 \cdot d_{\mathrm{min}}$ can be safely rejected. This implies that finding a tighter $d_{\mathrm{min}}$ in the early stages allows for the dismissal of more potential points.

Since the ED is a metric, the above equation applies, meaning any center $\vv{c}$ satisfying $\ED(\vv{c},\vv{c}_{\mathrm{best}}) > 2 \cdot \ED_{\mathrm{min}}$ can be discarded as nearest center. However, as the SED is not a metric, the TIE cannot be directly applied. Nonetheless, by squaring the aforementioned equation, given the non-negativity property of the ED, the following can be derived:
\begin{equation} \label{eq:TIEFilterSED}
    \SED(\vv{c},\vv{c}_{\mathrm{best}}) > 4 \cdot \SED_{\mathrm{min}} 
\end{equation}

This adjustment allows the use of the SED in a similar context.

\subsection{Norm-based filters}
\label{sec:norm-based-filters}

Given a point $\vv{p}$ and a center $\vv{c}$ in a $d$-dimensional space, the following two inequalities can be derived from the TIE:
\begin{enumerate}
    \item $d(\vv{O},\vv{p}) \leq d(\vv{O},\vv{c}) + d(\vv{c},\vv{p})$
    \item $d(\vv{O},\vv{c}) \leq d(\vv{O},\vv{p}) + d(\vv{p},\vv{c})$
\end{enumerate}
Hereby, $\vv{O}$ is the origin in $d$ dimensions. Given that $d$ \emph{must} be a metric to satisfy the TIE and that the ED between the origin $\vv{O}$ and any given point is equal to the norm of the point, we can express the previous equations as follows:
\begin{enumerate}
    \item $\norm{\vv{p}} \leq \norm{\vv{c}} + \ED(\vv{c},\vv{p})$
    \item $\norm{\vv{c}} \leq \norm{\vv{p}} + \ED(\vv{p},\vv{c})$
\end{enumerate}
Thus, knowing that $\norm{\vv{p}} -  \norm{\vv{c}} \leq \ED(\vv{p},\vv{c})$ and 
$ \norm{\vv{c}} - \norm{\vv{p}}\leq \ED(\vv{p},\vv{c})$, we obtain the following equation by combining them:
\begin{equation}
\label{eq:normAndED}
     \left | \;   \norm{\vv{c}} - \norm{\vv{p}} \; \right | \leq \ED(\vv{p},\vv{c}) \enspace,
\end{equation}
i.e., the difference in norm between point $\vv{p}$ and center $\vv{c}$ is constrained by their ED. Suppose we start with an initial center $\vv{c}_{\mathrm{best}}$ which we suppose is the closest one to point $\vv{p}$, and let $d_{\mathrm{min}}$ be their distance, i.e., $d_{\mathrm{min}} = \ED(\vv{p},\vv{c}_{\mathrm{best}})$. Then, any potential best center $\vv{c}$ must be closer  to $\vv{p}$ than $\vv{c}_{\mathrm{best}}$, i.e., $\ED(\vv{p},\vv{c}) \leq d_{min}$. Then, by Equation~\ref{eq:normAndED}, we obtain:
\begin{equation*}
    \left | \;   \norm{\vv{c}}- \norm{\vv{p}} \; \right | \leq \ED(\vv{p},\vv{c}) < d_{\mathrm{min}} \enspace,
\end{equation*}
which leads us to exclude any center $\vv{c}$ that fulfills:
\begin{equation}
    \left | \;   \norm{\vv{c}} - \norm{\vv{p}} \; \right | \geq d_{\mathrm{min}}
\end{equation}
This filter can also be applied using the SED by squaring both sides of Equation~\ref{eq:normAndED}, given that they are both positive, thus obtaining:
\begin{equation}
\label{eq:normAndSED}
     (\norm{\vv{c}} - \norm{\vv{p}})^2 \leq \SED(\vv{p},\vv{c})
\end{equation}

This norm-based filtering approach, as well as other similar methods, have been employed in the \Km literature \cite{Hamerly2014, Newling2016, Xia2020}, and have also found applications in closely related fields such as Vector Quantization \cite{KuangShyrWu2000} and Color Quantization \cite{Huang2021}.

\section{Accelerating \Kpp }

In this section, we first introduce a detailed description of the standard \Kpp algorithm. Then, we introduce an exact acceleration of the algorithm using the TIE along with a two-step sampling procedure, and further refine it using the previously introduced norm filters. 

\subsection{Standard \Kpp}

\begin{algorithm}[!htbp]
\caption{\Kpp}
\label{algo:KmeansPlusPlus}
\AlgoDisplayBlockMarkers\SetAlgoBlockMarkers{}{end}
\SetAlgoNoEnd
\DontPrintSemicolon
\Input{
points: $\mathcal{X} =\{\vv{x}_1, ..., \vv{x}_n\} \in \mathbb{R}^d$\;
\hspace{1.4cm} number of clusters: $k \in \mathbb{N}_{> 0}$\;
}
\BlankLine
    $\mathcal{C} \leftarrow \emptyset$\; 
    $c_{new} \leftarrow$ select $\vv{x} \in \mathcal{X}$ at random\;
    $\mathcal{C} \leftarrow \mathcal{C} \cup c_{new}$\;
    \While{$|\mathcal{C}| < k$} {
        $w_i \leftarrow \min_{c \in \mathcal{C}} \SED(\vv{x}_i, \vv{c})$ for all $x_i \in \mathcal{X}$\;
        $c_{new} \leftarrow$ select $\vv{x_i} \in \mathcal{X}$  with probability  $p_i = w_i / \sum_{j=1}^n w_j$\;
        $\mathcal{C} \leftarrow \mathcal{C} \cup c_{new}$\;
    }
    \Return $\mathcal{C}$
\BlankLine
\Output{
centers: $\mathcal{C} =\{\vv{c}_1, ..., \vv{c}_k\} \in \mathbb{R}^d$\;
}
\end{algorithm}

Algorithm~\ref{algo:KmeansPlusPlus} outlines the pseudo-code for the standard \Kpp algorithm. The algorithm starts by selecting a random point as the first center. Then, subsequent centers are selected using a ``roulette wheel selection'' mechanism (line $6$). Let $w_i$ be the weight assigned to each point $\vv{x}_i \in \mathcal{X}$, where $w_i$ is equal to the SED of the corresponding point to its closest center. Then, roulette wheel selection operates by assigning each point $\vv{x}_i$ a selection probability $p_i$ proportional to its weight $w_i$. More specifically, it selects point $\vv{x}_i$ with probability $p_i = w_i / \sum_{j=1}^n w_j$, i.e., it performs a weighed probability selection. Thus, points with a higher weight are more likely to be selected as centers. This is called $D^2$ sampling.

To this end, a random number $r$ is drawn from a uniform distribution between $0$ and the total sum of weights. We then iterate through the weighted points cumulatively, selecting the point $\vv{x}_i$ where the cumulative sum of weights just exceeds $r$. While this process typically requires linear time, $\mathcal{O}(n)$, due to the need to traverse through the points until the appropriate one is selected, it could potentially be optimized by pre-calculating cumulative weights and applying binary search. However, this optimization only yields time savings in scenarios where multiple points are chosen without altering the probability distributions, as it initially requires $\mathcal{O}(n)$ time to compute the cumulative weights but allows us to get the subsequent points in logarithmic $\mathcal{O}(\log n)$ time. However, in this scenario, given that the weights are updated after each center selection, and only one point is chosen per iteration, the potential gains from employing binary search are to be neglected.

Upon selecting a new center, the distance between each point and its closest center must be recalculated since the new center might become the closest to a subset of points (line 5). Therefore, the weight of a point is updated as $w_i = \min_{c \in C} \SED(\vv{x}_i, \vv{c})$. Although the straightforward procedure would entail checking all centers for each point---leading to $\mathcal{O}(kn)$ complexity per iteration and an overall complexity of $\mathcal{O}(k^2n)$---one can optimize by comparing each point only with the newly selected center and adjusting the closest center accordingly, using the fact that the closest center prior to the introduction of the new one remains the nearest among all predecessors. This optimization maintains the algorithm's runtime at $\mathcal{O}(kn)$. Then, this process is repeated until all of the centers have been selected. Note that the space complexity of this algorithm is $\mathcal{O}(n)$.

\subsection{Using the TIE}

Although the time complexity of the \Kpp algorithm is linear with respect to both the number of data points and clusters, it may become impractical for handling large datasets. As previously stated, the high computational cost comes from both phases: the calculation of the distances between each point and its closest center and the sampling procedure, as both need to iterate over the whole dataset in the worst-case scenario.

As previously stated, the TIE has been effectively employed to accelerate both the \Km \cite{Elkan2003, Hamerly2010} and \Kpp \cite{Xu2014, Raff2021} algorithms. In this work, we detail the application of TIE to the exact \Kpp algorithm. This is done in a similar way as in the recently introduced Ball \Km \cite{Xia2020}. Our application uses the SED instead of the ED and adapts the sampling procedure to this framework using a two-step procedure. The pseudo-code of the procedure is presented in Algorithm~\ref{algo:KmeansPlusPlusBall}.

\begin{algorithm}[htbp]
\caption{Accelerated \Kpp}
\label{algo:KmeansPlusPlusBall}
\AlgoDisplayBlockMarkers\SetAlgoBlockMarkers{}{end}
\SetAlgoNoEnd
\DontPrintSemicolon
\Input{$\mathcal{X} =\{\vv{x}_1, ..., \vv{x}_n\} \in \mathbb{R}^d$ (points), $k \in \mathbb{N}_{> 0}$ (\# of clusters)
}
\BlankLine
    $\mathcal{C} \leftarrow \emptyset$\; 
    $\vv{c}_{\mathrm{new}} \leftarrow$ select $\vv{x} \in \mathcal{X}$ at random \tcp*[r]{Select initial center}
    $\mathcal{C} \leftarrow \mathcal{C} \cup \{\vv{c}_{\mathrm{new}}\}$\;
    $\mathcal{P}_{\mathrm{new}} \leftarrow \mathcal{X}$ \tcp*[r]{Initial cluster contains all points}
    $w_i \leftarrow \SED(\vv{x}_i,\vv{c}_{\mathrm{new}})$ for all $\vv{x}_i \in \mathcal{X}$\;
    $r_{\mathrm{new}} \leftarrow \max\{w_i \mid \vv{x}_i \in \mathcal{P}_{\mathrm{new}}\} $\;
    $s_{\mathrm{new}} \leftarrow \sum_{\vv{x}_i \in \mathcal{P}_{\mathrm{new}}}w_i $\;
    \While{$|\mathcal{C}| < k$} {
        \tcp*[l]{Two-step sampling}
        $\mathcal{P}_{\mathrm{new}} \leftarrow $ select $\mathcal{P}_j$ with probability $p_j = s_j / \sum_{\vv{c}_l \in \mathcal{C}} s_l$\;
        $\vv{c}_{\mathrm{new}} \leftarrow $ select $\vv{x}_i \in \mathcal{P}_{new}$ with probability $p_i = w_i / s_{\mathrm{new}}$\;
        $\mathcal{C} \leftarrow \mathcal{C} \cup \{\vv{c}_{\mathrm{new}}\}$\;
         \For{$\vv{c}_j \in \mathcal{C} \setminus \{\vv{c}_{\mathrm{new}}\}$} {
          $d^{\mathrm{new}}_j \leftarrow \SED(\vv{c}_{\mathrm{new}},\vv{c}_{j})$ \;
          \If (\tcp*[f]{Filter 1}) {$4 \cdot r_j > d^{\mathrm{new}}_j $} {
            \For{$\vv{x}_i \in \mathcal{P}_j$} {
              \If (\tcp*[f]{Filter 2}) {$4 \cdot w_i > d^{\mathrm{new}}_j $} {
                $d_{\mathrm{new}} = \SED(\vv{x}_i,\vv{c}_{\mathrm{new}})$ \tcp*[r]{Calculate distance}
                \If {$w_i > d_{\mathrm{new}} $} {
                  $w_i = d_{\mathrm{new}}$\;
                  $\mathcal{P}_j \leftarrow \mathcal{P}_j \setminus \{\vv{x}_i\}$ \tcp*[r]{Remove from previous cluster}
                  $\mathcal{P}_{\mathrm{new}} \leftarrow \mathcal{P}_{\mathrm{new}} \cup \{\vv{x}_i$\} \tcp*[r]{Add to new cluster}
                }
              }  
            }
            $r_{j} \leftarrow \max\{w_i \mid \vv{x}_i \in \mathcal{P}_{j}\} $\;
    $s_{j} \leftarrow \sum_{\vv{x}_i \in \mathcal{P}_{j}} w_i\} $\;
          }
           
        }
         $r_{\mathrm{new}} \leftarrow \max\{w_i \mid \vv{x}_i \in \mathcal{P}_{\mathrm{new}}\} $\;
    $s_{\mathrm{new}} \leftarrow \sum_{\vv{x}_i \in \mathcal{P}_{\mathrm{new}}}w_i\} $\;
    }
    \Return $\mathcal{C}$
\BlankLine
\Output{
centers: $\mathcal{C} =\{\vv{c}_1, ..., \vv{c}_k\} \in \mathbb{R}^d$\;
}
\end{algorithm}

\subsubsection{First acceleration action}

The first acceleration action (named as \emph{Filter~1} at line~15 of Algorithm~\ref{algo:KmeansPlusPlusBall}) involves applying the TIE to bypass the calculation of distances between each point and a newly introduced center. Let $a(i)$ denote the center to which the point $\vv{x}_i$ is assigned, i.e., $a(i) = j \iff \arg \min_{\vv{c} \in \mathcal{C}} d(\vv{x}_i,\vv{c}) = \vv{c}_j$. Then, we denote cluster $j$ as $\mathcal{P}_j \subseteq \mathcal{X}$, where $\mathcal{P}_j$ denotes the set of points assigned to it, i.e., $\mathcal{P}_j = \{x_i \in \mathcal{X} \mid a(i) = j\}$, with  $\vv{c}_j$ representing the respective cluster center. Hence,  points are categorized based on their current assigned cluster.

For every cluster $j$, we maintain the maximum distance from its center $\vv{c}_j$ to any of its assigned points, denoted by $r_j$. This  maximum distance indicates that all points belonging to cluster $j$ are enclosed within a hyper-sphere of radius $r_j$ centered at $\vv{c}_j$. By the TIE rule previously introduced in Equation~\ref{eq:TIEFilter}, if the distance between a newly introduced center $\vv{c}_{\mathrm{new}}$ and the existing cluster center $\vv{c}_j$ exceeds twice the radius of that cluster's hyper-sphere, then it is guaranteed that no point within cluster $j$ is closer to $\vv{c}_{\mathrm{new}}$ than to $\vv{c}_j$. This is because, by definition, the distances from all points in the cluster to its center are less than or equal to $r_j$. As previously stated, the SED can be used instead, which is less computationally expensive. Hence, we can discard any cluster $j$ for which the following condition holds true:
\begin{equation}
    \SED(\vv{c}_j,\vv{c}_{new}) \geq 4 \cdot r_j \enspace,
\end{equation}
where, in this case, $r_j = \max \{\SED(\vv{x}_i,\vv{c}_j) \mid \vv{x}_i \in \mathcal{P}_j\}$.

As previously outlined, the first center $c_1$ is selected uniformly at random among the set of points. Consequently, all of the points initially belong to the first cluster ($\mathcal{P}_1$), with $a(i) = 1$ for all $\vv{x}_i \in \mathcal{X}$. Moreover, the radius $r_1$ is set to the maximum distance between any point and $\vv{c}_1$, i.e., $r_1 = \max \{\SED(\vv{x},\vv{c}_1) \mid \vv{x} \in \mathcal{X}\}$. This initialization can be found at lines 1 to 7.

Upon the selection of a new center $\vv{c}_{\mathrm{new}}$, we can bypass the comparison with all points of a cluster $j$ by using the previous equation, which implies that the new center $\vv{c}_{\mathrm{new}}$ is too distant from $\vv{c}_{j}$ to be nearer to its assigned points, thereby saving significant computation time. 

On the other hand, if this initial filter fails, we must iterate over all points within the cluster. Yet, the TIE can further be used to avoid the direct distance computations between a point $\vv{x}$ and the new center $\vv{c}_{\mathrm{new}}$ by using Equation~\ref{eq:TIEFilterSED} to forego the explicit distance computation. This is indicated as \emph{Filter 2} in the algorithm (line 17). If this second filter fails, then the respective distance must be calculated.

Finally, when a point $\vv{x}$ is deemed to be closer to the new center $\vv{c}_{\mathrm{new}}$ than its previously assigned one, it is removed from its previous cluster (line 21) and incorporated to the new cluster (line 22), where the radius $r_{new}$ is updated accordingly (lines 26 and 30). This process, thus, ensures accurate tracking of cluster distances and membership.

When a point $\vv{x}_i$ is removed from a cluster $j$ and $d(\vv{x}_i,\vv{c}_j) = r_j$, there is no efficient way of updating $r_j$ without revisiting all of its assigned points to find the new maximum value. This is also true when the new center is removed from its previous assigned cluster at each iteration. Nonetheless, $r_j$ requires modification only when all points must be reassessed due to the TIE filter failing. First, when a new center $c_{new}$ is removed from its previously assigned cluster, then all points from its previous cluster must be visited because the new center lies within that previous maximum radius $r_j$. Similarly, the other update of $r_j$ happens when points are removed from a cluster, which coincides with the TIE filter failure. Consequently, in both cases, as there is an imperative to examine each point within the affected cluster, the new $r_j$ can be determined by tracking the maximal distance from the cluster center to each remaining point, provided those points retain their original cluster assignment.

Implementing this technique, however, incurs a computational cost at each iteration, primarily due to the necessity of calculating the distances between the newly introduced center, $c_{new}$, and all existing cluster centers. Nonetheless, given that the number of clusters $k$ is significantly lower than the number of data points $n$ in most applications,  the impact of this overhead is rather negligible compared to the benefits derived from this approach. However, we will later discuss methods to potentially avoid part of these center distance calculations.

However, notice that the first acceleration action preserves the overall complexity of the respective algorithm phase at $\mathcal{O}(n)$. This is because, in the worst case, we might need to evaluate all points for potential reassignment to the new center. Nevertheless, it notably reduces the computational burden, particularly in the later stages of the algorithm and when a large number of clusters ($k$) is considered. 

\subsubsection{Second acceleration action}

The second enhancement concerns the roulette wheel selection procedure. In the worst case, performing roulette wheel selection requires iterating over the entire dataset. However, by grouping the points by the cluster to which they belong, we can streamline this process using the following two-step procedure. First, we can determine the cluster from which the next center will be selected by applying roulette wheel selection based on the sum of the weights of the points in each cluster. In the second phase, we only need to apply roulette wheel selection to the set of points belonging to the selected cluster.

As previously stated, the probability of each point being selected is proportional to the SED with respect to its assigned center. Rather than executing roulette wheel selection on the set of all points, the following is done. Let $s_j$ be the sum of the weights of all of the points in cluster $j$, i.e., $s_j = \sum_{\vv{x}_i \in \mathcal{P}_j} w_i$. Recall that $w_i = \SED(\vv{x}_i,\vv{c}_j)$. Then, we proceed as follows. First, the cluster from which the next center will be drawn is selected by roulette wheel selection with respect to the cluster weights $s_j$ (line~10). Thus, the probability of selecting a cluster $j$ is $p_j = s_j \ \sum_{l \in k} s_l$. Then, once the cluster is selected, we select a point from that cluster also by roulette wheel selection over the points of the selected cluster, knowing that their sum is equal to $s_j$ (line 11). Note that by doing this in two steps, we do not change the probabilities of each point being selected, and the procedure is equivalent to the standard sampling procedure. As the weights of the points of the clusters that we do not update remain the same, this approach circumvents the need to recalculate these sums at every step. Consequently, the next center is determined not through the conventional roulette wheel selection technique but via an adapted procedure that leverages these segmented probability sums. 

This adjustment reduces the expected complexity from $\mathcal{O}(n)$ to $\mathcal{O}(k+n/k)$, considering that, on average, each cluster contains $n/k$ points. In the worst-case scenario, i.e., when all points belong to the selected cluster, the complexity rises to $\mathcal{O}(k+n)$, though such scenarios are rare. However, since $k$ is assumed to be substantially smaller than $n$ ($k \ll n$), the impact on performance in this extreme scenario remains comparable to that of the traditional method.

Note that this procedure can even be further optimized. As outlined in the preceding section, executing the two-step roulette wheel selection procedure can achieve logarithmic complexity. Nevertheless, an initial complete pass through the whole dataset is required. In the context of the conventional \Kpp algorithm, this approach proves no benefit since each center addition requires iterating over the entire dataset. However, in the modified approach, where many clusters are not iterated over, it becomes feasible to compute these cumulative sums each time a cluster is visited. Then, these pre-calculated sums remain valid for subsequent iterations as long as the cluster remains unchanged, enabling the application of binary search on the cumulative weights within each cluster. Additionally, given that a review of all clusters is necessary at least once per iteration to check for the filters, it is also viable to compute the cumulative sums over the cluster weights during this phase.  Consequently, the roulette wheel method---or $D^2$ sampling---can be executed in logarithmic time in this two-step approach, further accelerating the whole process.

\subsection{Using additional geometric information}

The TIE filter proves to be particularly effective at the later stages of the process. As additional centers are incorporated, the radius associated with each cluster tends to diminish, thus increasing the likelihood of excluding other clusters from consideration. Nonetheless, this procedure can be further refined, especially in the algorithm's early stages, where clusters are larger, which can limit the computational savings achievable through this filter.

\begin{figure}[htb]
  \centering
  {\includegraphics[width=0.75\textwidth]{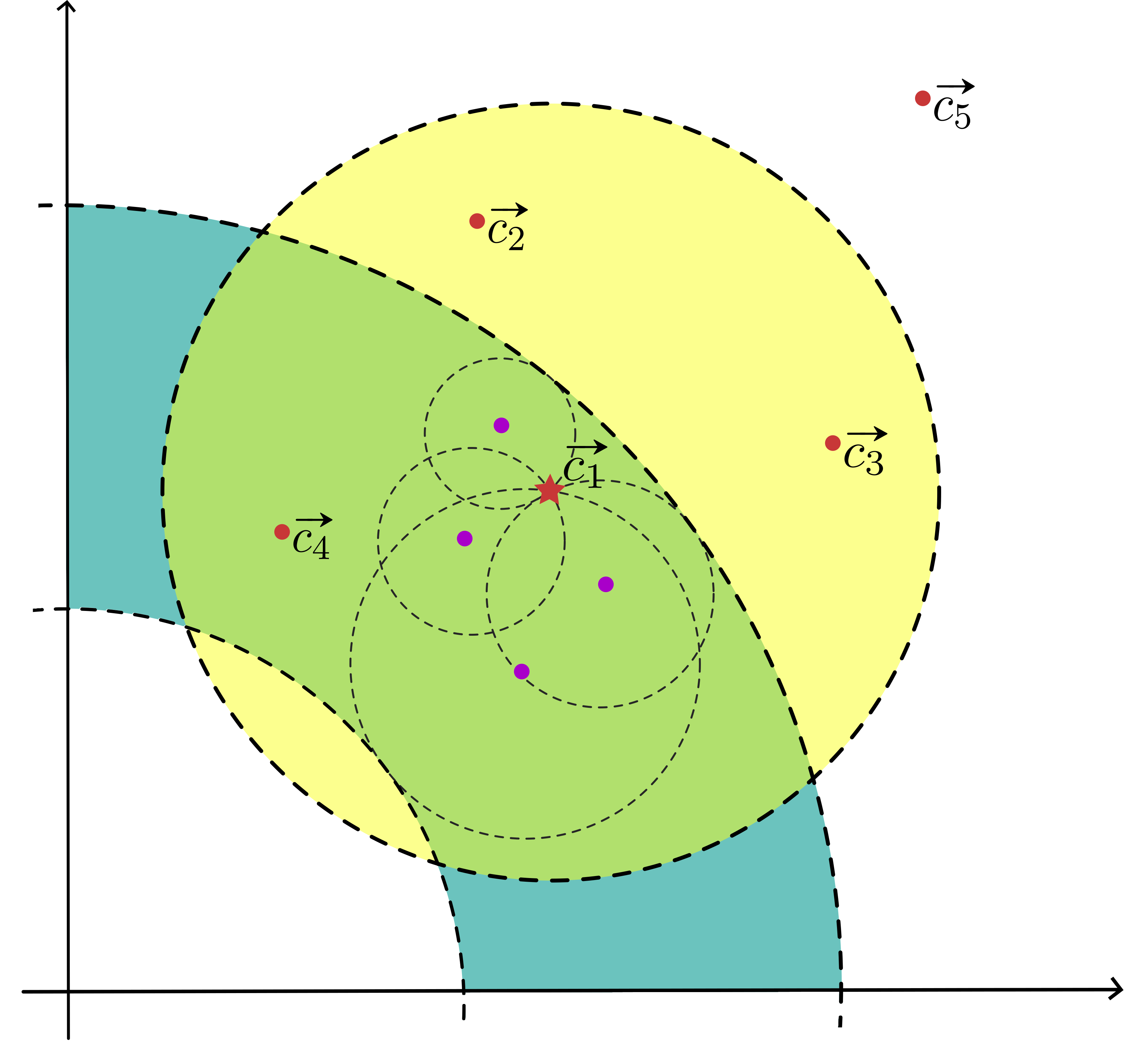}}
  \caption{Clustering example in two dimensions. \\ The data points are depicted in purple and the centers in red, with the currently assigned center marked by a star shape. The area highlighted in yellow represents twice the radius of the cluster formed by these points (assuming ED). 
  Around each data point, there is a dashed circle with a radius equal to the ED to the point's assigned center. The blue-shaded area delimits the space between the lower and upper bounds for centers to be considered.}
  \label{fig:norm_example}
\end{figure}

Figure~\ref{fig:norm_example} shows an example of a clustering in a two-dimensional space. The data points are depicted in purple and the centers in red, with the currently assigned center marked by a star shape. The area highlighted in yellow represents twice the radius of the cluster formed by these points, assuming we are using the ED for simplicity. Thus, any center falling within this yellow area would not pass the TIE filter, and therefore, a review of all points in the cluster to determine if they should be reassigned to the newly introduced center would be needed. In this case, it is apparent that centers $\vv{c}_2$ and $\vv{c}_3$, although lying within this area, cannot be the closest center to any of the points due to the points' spatial distribution. This observation suggests that additional geometric insights could further exclude potential centers from consideration. To this end, we integrate the norm filters mentioned earlier (see Section~\ref{sec:norm-based-filters}) to refine the search space further. Recall that the norm difference between a point and a center is constrained by their Euclidean Distance. Therefore, if the norm difference between a point and the newly added center exceeds the distance between the point and its assigned cluster, the point cannot be part of the newly added cluster.

To leverage this property, we further divide each cluster $\mathcal{P}_j$ into two partitions: the lower partition $\mathcal{L}_j$ and the upper partition $\mathcal{U}_j$. The lower partition of a cluster is formed by the points whose norm is less than or equal to the center norm, i.e., $\mathcal{L}_j = \{\vv{x}_i \in \mathcal{P}_j \mid \norm{\vv{x}_i} \leq \norm{\vv{c}_j}\}$. Conversely, the upper partition comprises points whose norm is greater than the center norm, i.e., $\mathcal{U}_j = \{\vv{x}_i \in \mathcal{P}_j \mid \norm{\vv{x}_i} > \norm{\vv{c}_j}\}$. Note that $\mathcal{L}_j \cup \mathcal{U}_j = \mathcal{P}_j$.

Next, we define the lower bound $l(\vv{x}_i)$ of a point $\vv{x}_i$ to be equal to its norm minus the ED to its assigned center, i.e., $l(\vv{x}_i) = \norm{\vv{x}_i} - \ED(\vv{x}_i,\vv{c}_a(i))$. Conversely, we define the upper bound $u(\vv{x}_i)$ of a point $\vv{x}_i$ to be equal to the sum of both variables, i.e., $u(\vv{x}_i) = \norm{\vv{x}_i} + \ED(\vv{x}_i,\vv{c}_a(i))$. Note that, given the previously defined property of the norms (Equation~\ref{eq:normAndED}), it follows that if a new center does not fall within these bounds, it cannot be the nearest center to the point. Thus, we can avoid computing the distance between them. For this purpose, we can use Equation~\ref{eq:TIEFilterSED} when working with the SED.

Similarly, we extend these upper and lower bound definitions for partitions. In this case, the lower bound of a partition is set to be the minimum lower bound among all points in the partition, while the upper bound is set to be the maximum upper bound of the points. More specifically, given a partition $\mathcal{L}_j$, we define $l(\mathcal{L}_j) = \min\{l(\vv{x}_i) \mid \vv{x}_i \in \mathcal{L}_j\}$ and $u(\mathcal{L}_j) = \max\{u(\vv{x}_i) \min \vv{x}_i \in \mathcal{L}_j\}$.  Following this rationale, any new center that lies outside these bounds for a partition cannot be the nearest to any points within that partition. Note that the same procedure is similarly applicable to the upper partition $\mathcal{U}_j$.

Going back to Figure~\ref{fig:norm_example}, note that around each data point, there is a dashed circle with a radius equal to the ED to the point's assigned center. On the basis of these dashed circles, the lower and upper bounds of the partition formed by these points are computed. The area between these lower and upper bounds is shown in Figure~\ref{fig:norm_example} as the blue-shaded area. Any center that does not fall within this area can safely be dismissed from consideration. In the case of Figure~\ref{fig:norm_example}, centers $\vv{c}_2$ and $\vv{c}_3$ can therefore be discarded using this norm filter. 


However, implementing this strategy requires some additional computations. First, the norms for all points and centers need to be calculated, which can be efficiently pre-computed at the start of the algorithms' execution since they remain constant. Moreover, the ED is required for the calculation of the upper and lower bounds of each point, as the SED does not fulfill the TIE. Nevertheless, the SED remains applicable for calculating distances between clusters in each iteration.

\section{Experimental Evaluation}

\subsection{Instances}

To test the efficiency of our proposed approach, we use real-world instances. Table~\ref{tab:instances} shows the list of instances, along with (1) their size $n$ in terms of the number of points and (2) the dimension $d$ of the points. Instances have been pre-processed by removing data points with missing values. They are categorized into low-dimensional instances (first 12 table rows) and high-dimensional instances (last 9 table rows), defined as those with dimensions less than or greater than 16, respectively. Within these groups, instances are ordered by increasing size. Moreover, all available data points were merged into one set for those test instances with data points separated into training, validation, and testing sets. Note that in those cases in which the instance dimension ($d$) shown in Table~\ref{tab:instances} does not coincide with the dimension information given in the original source, we have reduced the original dimensionality by removing features irrelevant to our objectives, such as identifiers, dates, times, labels, classes or tags. Some specific features that showed weak correlations or relevance to the main features under consideration were removed from certain instances. All test instances can be obtained from the corresponding author on request.

\begin{table}[!t]
\caption{List of real-world instances used for the experimental evaluation}
\label{tab:instances}
\centering
\resizebox{\columnwidth}{!}{
\renewcommand{\arraystretch}{1.3}
\begin{tabular}{lrrr}

    \hline
    \multirow{2}{*}{Instance} & \multirow{2}{*}{$n$}  &  \multirow{2}{*}{$d$} & \% norm \\
    & & & variance\\
    \hline
    MAGIC Gamma Telescope (\texttt{MGT})  \cite{Heck1998, Bock2004-Dataset} & $19,020$ & $10$  & $50.00$\\
    \hline
    Corel Image Features - Color Moments (\texttt{CIF-C}) \cite{Ortega1998-Dataset} & $68,040$ & $9$ & $11.49$ \\
    \hline
    Corel Image Features - Co-occurrence Texture (\texttt{CIF-T}) \cite{Ortega1998-Dataset} & $68,040$ &  $16$ & $48.06$\\
    \hline
    Query Analytics Workloads - Range Queries Aggregates (\texttt{RQ}) \cite{Anagnostopoulos2018-Dataset} & $200,000$ & $7$ & $2.60$ \\ 
    \hline
    Skin Segmentation Skin-NonSkin (\texttt{S-NS}) \cite{Bhatt2009-Dataset} & $245,057$  & $3$ & $75.45$\\
    \hline
    3D Road Network (\texttt{3DR}) \cite{Kaul2013-Dataset, Kaul2013} & $434,874$  &  $3$ & $22.63$ \\
    \hline
    COD-RNA (\texttt{RNA}) \cite{Uzilov2006}  & $488,565$ &  $6$ & $8.97$\\
    \hline
    Household Power Consumption (\texttt{HPC}) \cite{Hebrail2006-Dataset} & $2,049,280$  &  $7$  & $5.40$\\
    \hline
    HAR70$+$ Human Activity Recognition  (\texttt{HAR}) \cite{Logacjov2023-Dataset, Ustad2023} & $2,259,597$ & $6$ & $10.43$\\ 
    \hline
    Gas Sensor Array Dynamic Mixtures - CO (\texttt{GS-CO}) \cite{Fonollosa2015-Dataset, Fonollosa2015} & $4,208,262$ & $16$ & $85.12$\\
    \hline
    Gas Sensor Array Dynamic Mixtures - Methane (\texttt{GS-MET}) \cite{Fonollosa2015-Dataset, Fonollosa2015} & $4,178,505$ & $16$ & $56.38$\\
    \hline
    Yahoo! Webscope: R6A Today Module - Users (\texttt{YAH}) \cite{YahooWebscope} & $45,811,883$ & $5$ & $4.84$\\
    \hline
    \multicolumn{4}{c}{ } \\
    \hline
    Gas Sensor Array Drift (\texttt{GSAD}) \cite{Vergara2012-Dataset, Vergara2012, RodriguezLujan2014} & $13,910$ &  $128$ & $85.56$ \\
    \hline 
    KDD - Physics (\texttt{PHY}) \cite{Caruana2004} & $18,644$ & $78$ & $7.48$ \\ 
    \hline
    Crop (\texttt{CRP})  \cite{Tan2017, UCRArchive2018} & $24,000$ & $46$ & $52.92$\\ 
    \hline
    CIFAR-10 (\texttt{C-10}) \cite{Krizhevsky2009} & $60,000$  &  $3,072$ & $23.61$  \\  
    \hline
    CIFAR-100 (\texttt{C-100}) \cite{Krizhevsky2009} & $60,000$  &  $3,072$  & $28.08$ \\
    \hline
    MNIST Database of Handwritten Digits (\texttt{MNIST}) \cite{Lecun1998} & $70,000$ & $784$ & $5.51$ \\
    \hline
    KDD - Protein (\texttt{PTN}) \cite{Caruana2004}  & $285,409$ & $74$ & $85.12$\\
    \hline
    Million Song Dataset - Year Prediction (\texttt{YP}) \cite{Bertin-Mahieux2011, Bertin-Mahieux2011-Dataset} & $515,345$ & $90$ & $61.49$ \\
    \hline
    Supersymmetry (\texttt{SUSY}) \cite{Baldi2014,SUSY-Dataset} & $5,000,000$ & $18$ & $20.96$\\
\hline
\end{tabular}
}
\end{table}

\subsection{Performance evaluation}

The following three algorithm variants are considered in the experiments: (1) standard \Kpp, (2) accelerated \Kpp without norm filter (that is, only using the TIE filter), and (3) full accelerated \Kpp. Each algorithm variant was applied 10 times to each combination of a problem instance and a cluster number $k \in \{2^0 = 1, \ldots, 2^{12}=4096\}$. All results are shown in terms of mean values over the 10 algorithm applications. 

For the evaluation, we focused on the following key metrics, which we aim to optimize: (1) the portion of the dataset examined during the identification of the new closest center, (2) the portion of the dataset examined during the $D^2$ sampling phase, (3) the total number of distance computations performed between points and centers and (4) the total running time. 

The first three metrics serve as indicators of the algorithm's intrinsic efficiency, highlighting improvements brought by our optimized approach. These measures are particularly valuable as they remain unaffected by variables external to the algorithm, such as the computing environment or implementation specifics. While the primary objective of our approach is to reduce the total computational time, this aspect is inherently susceptible to these external variables. However, it is crucial to evaluate whether the theoretical gains in speed are nullified by any additional computational overhead introduced in the optimized variant. Thus, we aim to validate whether our accelerated version theoretically reduces computational demands and obtains significant gains in practical applications. Lastly, in subsequent experiments, we will show how the computation environment can affect the running time of the algorithm. \\

\begin{figure}[htbp]
  \centering
  {
  \includegraphics[width=\textwidth,height=0.95\textheight]{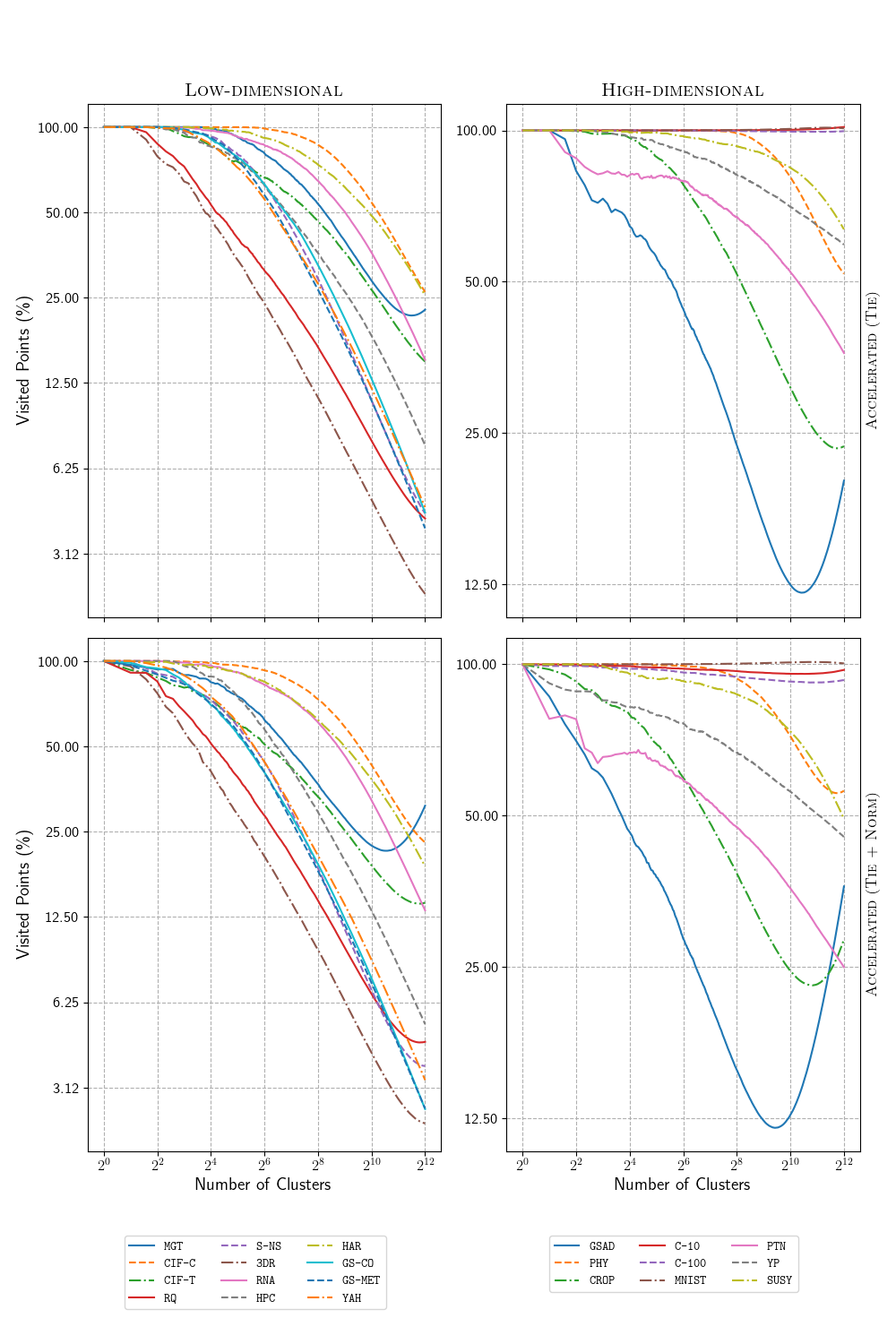}}
  \caption{Percentage of examined points 
  (in relation to the standard \Kpp) 
  for the accelerated \Kpp version using only the TIE filter (upper row), 
  and for the accelerated \Kpp version that also uses the additional norm filter
  (lower row).}
  \label{fig:visited}
\end{figure}

\begin{figure}[htbp]
  \centering
  {
  \includegraphics[width=\textwidth,height=0.95\textheight]{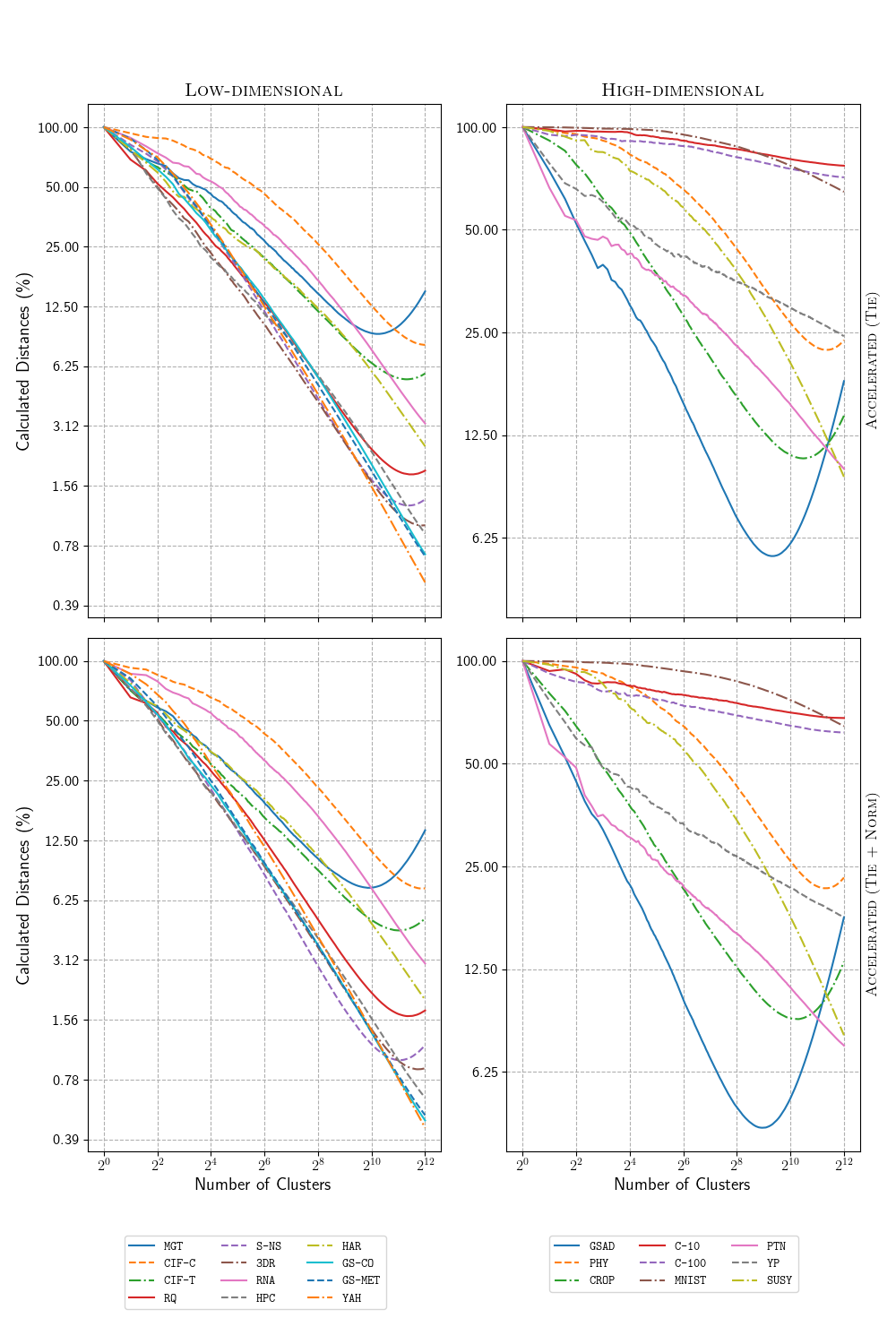}}
  \caption{Percentage of calculated distances 
  (in relation to the standard \Kpp)
  for the accelerated \Kpp version using only the TIE filter (upper row), 
  and for the accelerated \Kpp version that also uses the additional norm filter
  (lower row).}
  \label{fig:calculated}
\end{figure}

Figure~\ref{fig:visited} illustrates for all cluster numbers between $k=1$ and $k=n^{12}=4096$ the percentage of data points examined by our accelerated \Kpp in relation to the total number of data points examined by the standard \Kpp algorithm. Note that each curve in these graphics shows the evaluation of this percentage for exactly one of the datasets, that is, instances. For a clearer presentation of the results, the datasets are categorized into two types: low-dimensional and high-dimensional instances. Low-dimensional instances are those with a dimension ($d$) of 16 or less. In particular, the left-hand-side graphics contain the results for low-dimensional instances, while the right-hand-size graphics are for high-dimensional instances. The first row of graphics presents results for the accelerated \Kpp version using only the TIE filter, while the second row shows results for the accelerated \Kpp version that also uses the additional norm filter. To ensure fairness, we have counted the visited clusters as points examined in both accelerated versions (or partitions in the second). Note that Figure~\ref{fig:calculated} shows the results concerning the percentage of the calculated distances (with respect to the total number of calculated distances in standard \Kpp) in the same way as explained above. For the calculated distances, we have also included the pairwise center distances calculated at each iteration. In the case of the accelerated version with the additional norm filter, the number of norms calculated is also included. It is important to note that these calculations occur only at the first iteration. Both axes are in logarithmic scale to more clearly show the performance of the algorithm. 


\begin{figure}[htbp]
  \centering
  {\includegraphics[width=\textwidth]{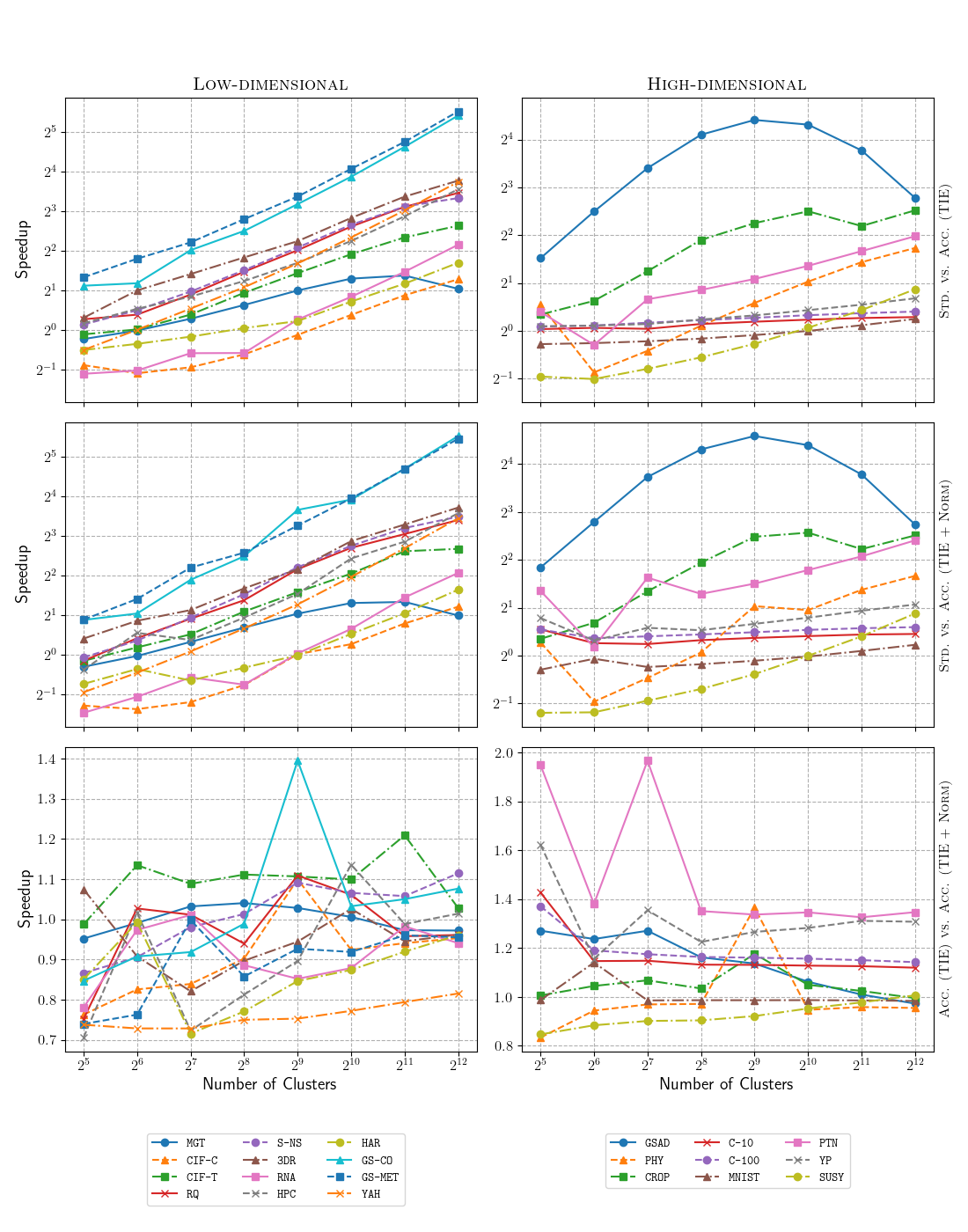}}
  \caption{Speedups of the accelerated \Kpp variants with the standard \Kpp algorithm (first and second row), and speedup of the full accelerated \Kpp variant with the \Kpp variant that does not use the norm filter (third row).}
  \label{fig:speedups}
\end{figure}

Furthermore, Figure~\ref{fig:speedups} depicts---in the same way as explained above---the speedup results of the accelerated \Kpp variants (with and without norm filter) with the standard \Kpp algorithm (first and second row of graphics), and the speedup results when comparing the full accelerated \Kpp variant with the \Kpp variant that does not use the norm filter. In this context, note that the speedup is defined as the mean time of the first algorithm divided by the mean time of the second. Therefore, the speedup indicates how much faster the second algorithm performs compared to the first.

\begin{figure}[!t]
  \centering
  {\includegraphics[width=\textwidth]{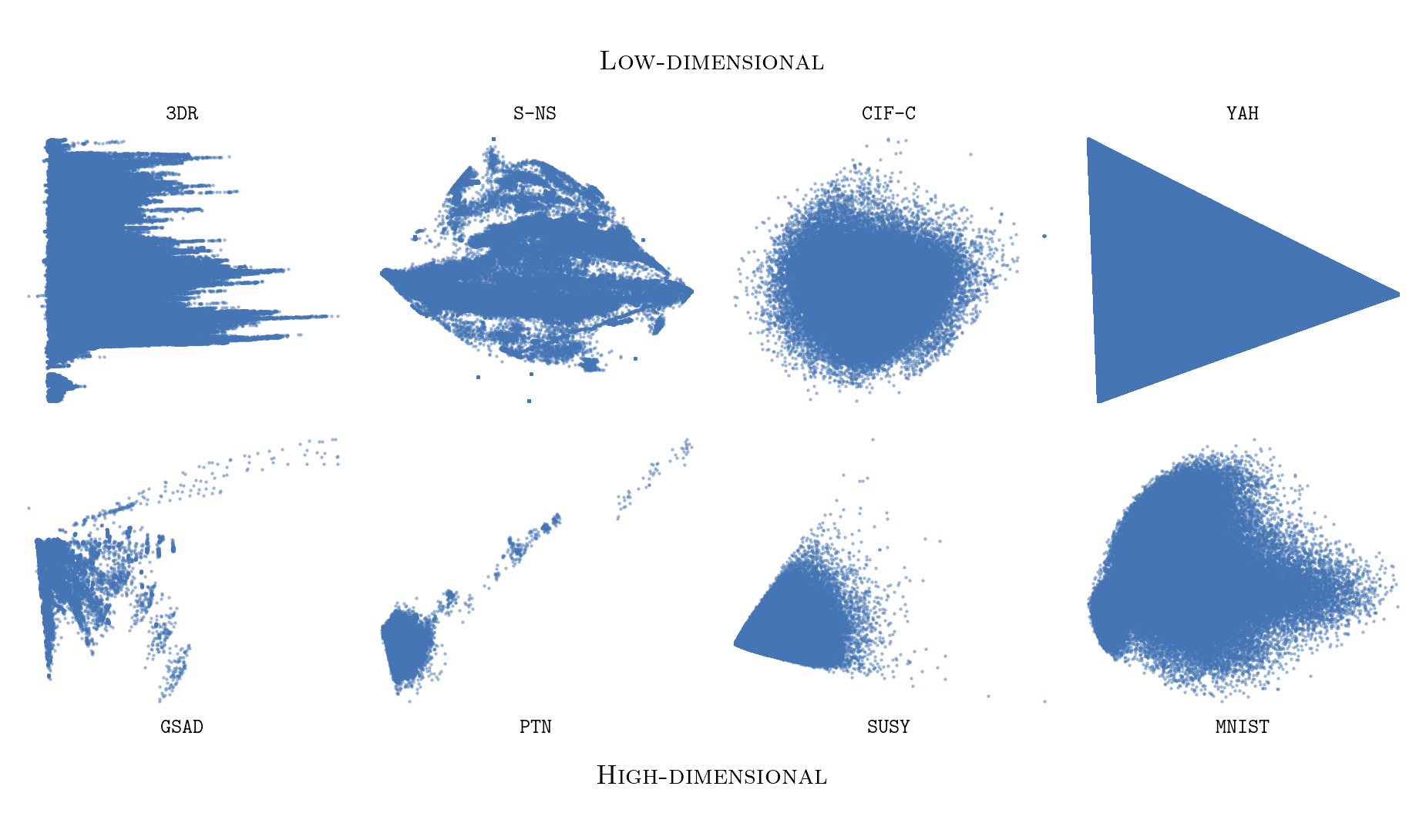}}
  \caption{Two-dimensional visualization of a subset of instances using PCA, 
  for low-dimensional (top row) and high-dimensional (bottom row) instances.}
  \label{fig:examples}
\end{figure}

Finally, Figure~\ref{fig:examples} presents a two-dimensional visualization of the datasets for a subset of low-dimensional (top row) and high-dimensional (bottom row) instances using Principal Component Analysis (PCA).\\

Before delving into the description and interpretation of the results, we would like to make some clarifying remarks. In some cases, the reduction in computational distance calculations does not seem to be accompanied by an improving algorithm speed. This phenomenon can primarily be attributed to the following two factors:
\begin{enumerate}
    \item First, the dimensionality of the data may significantly impact the speedup gains with respect to calculation savings. Specifically, reducing distance calculations in higher-dimensional spaces can result in more substantial time savings compared to similar reductions in lower-dimensional spaces  This is because the computational cost of calculating distances in high-dimensional spaces is inherently greater; thus, any reduction in these calculations can lead to more noticeable improvements in performance. For instance, this can be observed in instances like \texttt{GS-CO}/\texttt{MET} versus \texttt{3DR} in low dimensions and \texttt{CIFAR-10}/\texttt{100} versus \texttt{SUSY} in the case of high-dimensional instances.
    \item Second, the practical performance of accelerated algorithms is often influenced by various system-level factors, particularly the memory management. Accelerated algorithms tend to access memory in a less ordered manner than the standard algorithm variant, potentially affecting cache locality. This irregular memory access pattern is especially pronounced in the full accelerated \Kpp which is using the additional norm filter, due to the clusters being further divided into two partitions. Furthermore, processing high-dimensional data points is more memory-intensive, as fitting these points into the limited capacity of memory or cache allows for fewer points to be stored simultaneously. Consequently, if the algorithm needs to access more points, the overall memory performance can degrade. We will explore this more deeply later on.
\end{enumerate}

\subsubsection{Results of the accelerated \Kpp variant only using the TIE filter}

\noindent The following observations can be made:
\begin{itemize}
    \item The algorithm generally outperforms the standard \Kpp, achieving speedups of up to $2^5-2^6$ in some instances. It obtains better results in low-dimensional settings, which was to be expected given the curse of dimensionality. Overall, the efficacy of TIE tends to grow with an increasing number of clusters. As the \Kpp algorithm inherently produces well-separated clusters, the filtering mechanism of TIE becomes progressively more effective. Conversely, when a low number of clusters is considered, TIE loses efficacy due to the lower number of point visit savings identified by the TIE filter, combined with the additional overhead from computing extra pairwise distances.
    
    \item The spatial distribution of points within each dataset significantly influences the algorithm's performance. For instance, in some low-dimensional instances such as \texttt{CIF-C}, \texttt{CIF-T}, \texttt{RNA}, \texttt{HAR}, and \texttt{YAH}, the accelerated \Kpp hardly causes any improvement for low numbers of clusters.
    
    Visualization of these datasets in two dimensions using PCA reveals a lack of distinct separation between points. This is more noticeable in instances like \texttt{CIF-C} and \texttt{HAR}, where points are densely distributed around a central mass,  compared to slightly more dispersed structures in the other datasets. In contrast, instances like \texttt{YAH} exhibit a more uniform distribution across the visible cluster. Such uniformity generally enhances performance as the algorithm progresses and more centers are selected. Initially, the algorithm struggles due to the apparent absence of separation between points due to the reduced effectiveness of the TIE in this context. However, as the number of clusters increases, the formation of well-separated, balanced clusters augments the effectiveness of the TIE filter.
    
    Conversely, instances with a densely packed cluster around a central point often lead to imbalances in cluster formation, which reduces the filter's efficiency. In such cases, points concentrated at the center are more frequently selected, reducing the efficiency of the TIE filter by only bypassing the outer, less numerous points. While farther apart points may be more likely to be selected due to their distance from already chosen centers, their smaller number makes them less likely to be chosen, thus not sufficiently compensating for iterating over central points. Thus, cluster imbalance is detrimental; avoiding clusters with fewer points proves costly due to the increased overhead of calculating distances to the center, thus reducing the gain of avoiding visiting that cluster.

    \item As expected, high-dimensional instances present a more challenging environment, leading to diminished speedups due to the curse of dimensionality affecting the TIE. Again, datasets with well-separated clusters, like \texttt{GSAD} or \texttt{PTM}, cause an augmented usefulness of the TIE filter compared to those with dense, cloud-like formations such as \texttt{SUSY} or \texttt{MNIST}.

    \item Smaller instances such as \texttt{MGT} or \texttt{GSAD} exhibit a decrease in speedup, primarily attributed to the additional computations outweighing the benefits from reduced distance calculations. This issue becomes more pronounced with an increase in the number of centers; the standard \Kpp algorithm benefits from computing fewer distances with each iteration, whereas the accelerated version incurs additional overhead by calculating more pairwise center distances. The impact of this inefficiency is further magnified when the ratio of the number of clusters to the total number of points is relatively low.
\end{itemize}

\subsubsection{Results of the full accelerated \Kpp variant}

In general, the full accelerated \Kpp algorithm exhibits performance trends similar to those of the accelerated version using only the TIE, with some notable differences:
\begin{itemize}
    \item In lower-dimensional settings, particularly at smaller values of $k$, this version performs slightly worse than the TIE-only version. Concerning the \texttt{RNA} and \texttt{CIF-C} instances, for example, it initially underperforms but achieves comparable results to the TIE-only variant as the number of clusters increases. This is expected because---even though the norm filter effectively filters additional points not identified by the TIE filter at lower values of $k$---it introduces an additional overhead by initially having to calculate all the points' norms. Thus, this added complexity can be counterproductive in scenarios where the computational savings are not as high.
    \item Despite the norm filter reducing the number of calculations in almost all cases, this does not always translate into increased speedup for two primary reasons. Firstly, in some instances, the marginal savings do not significantly impact performance in lower dimensions as much as in higher dimensions, where reducing distance calculations typically yield more substantial gains. Secondly, this version demonstrates poorer data locality, an aspect that will be further explored in upcoming experiments. 
    \item In some instances, like \texttt{YAH}, the performance of the full accelerated \Kpp algorithm is notably inferior compared to the TIE-only version. This is partly due to the lower dimensionality of the dataset and the fact that the variance in norms is among the lowest in the dataset at $4.84$. Consequently, the norm filter proves ineffective, serving only to introduce unnecessary overhead.
    \item In other instances, such as \texttt{S-NS}, the norm-filtered version starts off underperforming but gradually recovers as the number of clusters increases. This recovery is attributed to the effectiveness of the norm filter in this context, where the norm variance is high at $75.45$. This is expected as the data points in \texttt{S-NS} are pixel values typically distributed within the RGB cube, making norms particularly useful for distinguishing points. While the TIE performs better at a higher number of clusters, the combined use of both the norm filter and TIE, in this case, enhances the overall effectiveness. It is worth noting that further subdividing the clusters into two partitions enhances the precision of the TIE, as each partition utilizes its own specific radius.
    \item In cases like \texttt{HAR} or \texttt{HPC}, the use of the norm filter is effective despite their low norm variance. This can be attributed firstly to the further partitioning of clusters, which allows the TIE filter to achieve greater precision. Additionally, since these instances are slightly higher in dimensionality compared to others in the lower-dimensional group, even minor computational savings have a higher impact on speedup due to the reduced need to calculate distances among points and centers. Initially, the radius proves more effective, but its advantage diminishes as more clusters are formed, at which point both filters demonstrate effectiveness.
    \item Concerning instances \texttt{CIF-C} and \texttt{CIF-T}, which are rather similar, the algorithm performs better for the texture instance (\texttt{CIF-T}) than for the color instance (\texttt{CIF-C}). This difference is expected given the variance in norms: \texttt{CIF-T} has a norm variance of $48.06$, significantly higher than \texttt{CIF-C}’s $11.49$, making the norm filter less effective in the latter. A parallel can be drawn with higher-dimensional datasets like \texttt{GS-CO} and \texttt{GS-MET}, where the former causes a higher improvement of the full accelerated \Kpp due to a higher norm variance of $85.12$ compared to $56.38$ for the latter. These examples clearly illustrate why the norm filter performs variably across different datasets; similar datasets show divergent results primarily due to differences in norm variance.
    \item In high-dimensional instances, the full accelerated \Kpp generally outperforms the TIE-only variant. This is expected because, as previously discussed, savings from reduced distance calculations have a more substantial impact in high-dimensional settings.
    \item The most notable speedup is observed in the \texttt{PTN} instance, which exhibits a high norm variance of $85.12$. This makes sense since many points are more effectively filtered by the norm filter compared to the radius filter, and the dataset itself has a relatively high dimension. In contrast, the \texttt{PHY} instance---despite having a similar dimension---shows a drastically lower norm variance of only $7.48$, making the norm filter far less effective. Additionally, because \texttt{PHY} is a smaller instance, the minimal savings do not compensate for the increase in the number of extra pairwise distance and norm calculations, which overshadow any benefits from reduced distance calculations. A similar situation occurs with the \texttt{GSAD} dataset. Although it has a high norm variance of $85.56$, making the norm filter highly effective initially, the speedup quickly diminishes due to the small size of the dataset. As mentioned earlier, the radius filter also performs well in this context due to the dataset's well-separated nature. Thus, combining both filters leads to significant savings in distance calculations.
    \item For the \texttt{C-10} and \texttt{C-100} instances, which are similar in nature, it is observed that \texttt{C-100}, having a higher norm variance than \texttt{C-10}, provokes a slightly better speedup. This indicates that the higher norm variance in \texttt{C-100} contributes to this improved performance despite their similarities.
    \item The \texttt{YP} dataset, which has a relatively high norm variance of $52.92$, is another example that demonstrates the effectiveness of the norm filter. In this case, this filter significantly outperforms the radius filter, as evidenced by the greater speedup observed in the TIE variant with the norm filter compared to the radius filter alone.
    \item Conversely, for the \texttt{SUSY} instance, which has a norm variance of $20.96$, the full accelerated \Kpp shows a poorer performance. This may be because it has one of the lowest dimensionalities within this group coupled with a medium variance, which might not be sufficient to leverage the norm filter's benefits effectively.
\end{itemize}

\subsection{Hardware-Related Performance Issues}

As previously stated, the theoretical gains expected from algorithm optimizations do not always go along with the performance improvements measured in practice. For example, while the additional norm filter generally reduces the number of computations and visited points, this does not consistently translate into speedups. Initially, one hypothesis was that both accelerated versions, especially the one with the norm filter, suffer from poor data locality compared to the standard \Kpp variant, which sequentially processes points. This deviation might lead to increased cache misses, adversely affecting the algorithm's performance. Additionally, as mentioned earlier, the implementation and the computing environment can significantly influence the practical performance (time) of the algorithms. In this context, note that the previously reported experiments were conducted in parallel on a computing cluster, whereby all applications for a single instance were sent simultaneously to the clusters' queue. This setup raises questions about whether such concurrency might skew the expected results.

To investigate these factors further, we conducted the following experiments: An algorithm is first run in isolation for a specific instance, meaning that only that specific algorithm/instance combination is executed on the system. Then, the same combination is run in $j$ concurrent jobs, with $j$ scaling up to 10. Each algorithm application is repeated 10 times. These tests are performed on a computer cluster equipped with two processors, each having 12 cores.

We decided to use the \texttt{3DR} instance for our analysis. First, low-dimensional data helps us better identify memory issues related to data access, as high-dimensional data could occupy almost all of the cache, causing cache misses due to fewer points being stored there rather than access patterns. Additionally, we observed that although the fully accelerated version visits slightly fewer points and performs fewer calculations than the TIE-only version, this does not result in an actual speedup. This discrepancy is what we aim to investigate.

During each run, we measure several performance metrics, including execution time, the percentage of level 1 cache misses, the percentage of last-level cache misses, and the number of instructions per cycle. These metrics will help us identify and analyze performance patterns across different contexts.

\begin{sidewaysfigure}[htbp]
  \centering
  \includegraphics[width=1.1\textwidth]{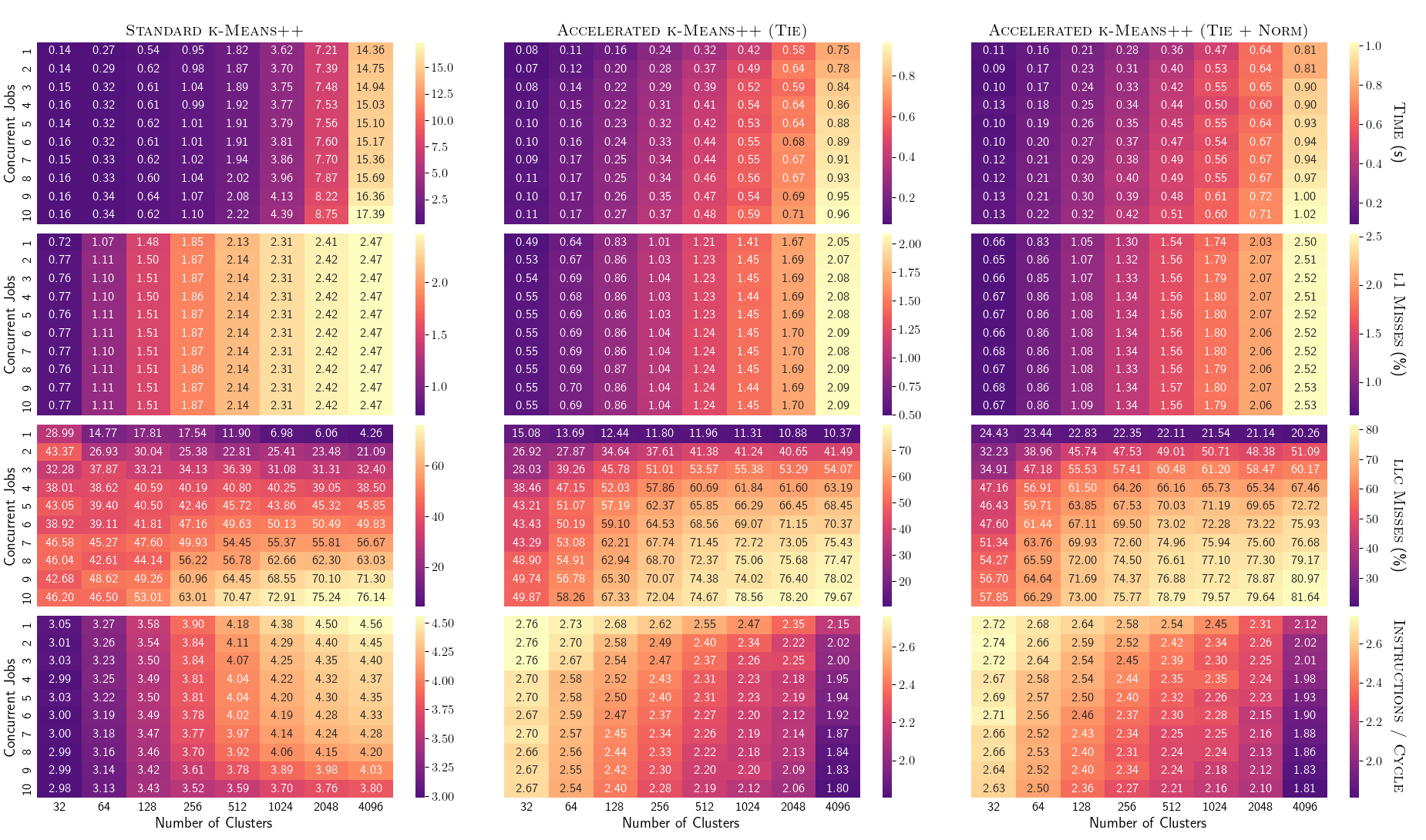}
  \caption{Heatmaps illustrating the time and memory performance of the standard \Kpp and the two accelerated algorithm variants under a varying number of concurrent jobs.}
  \label{fig:perf_heatmaps}
\end{sidewaysfigure}

Figure~\ref{fig:perf_heatmaps} presents the obtained results using heatmaps. Each heatmap's x-axis represents the number of clusters in increasing order, while the y-axis represents the number of concurrent jobs (from $1$ to $10$). Each cell in the heatmap represents the average result for each specific combination. Hereby, for each combination, the algorithm was applied 10 times to ensure reliability in the collected data. Each column indicates the algorithm used (the standard variant and the two accelerated algorithm variants, respectively), and each row corresponds to a specific metric being evaluated. Execution time is reported in seconds, cache misses are expressed as percentages (\%), and the last metric is the number of instructions per cycle (IPC).

\subsubsection{Organization of computer memory}

We briefly review how computer memory is structured to understand the performance results. Generally, a CPU comprises one or more cores, each capable of executing instructions. RAM (Random Access Memory) stores data and programs that the CPU is actively using. RAM is significantly faster at reading and writing data than hard drives or SSDs, essentially acting as the primary workspace for the processor, allowing for faster data access. Moreover, each processor includes caches, which are small, high-speed memory layers located within or adjacent to the CPU. These caches store frequently accessed data and instructions from slower memory sources, like RAM, to speed up retrieval times and, thus, CPU performance. Caches are organized into levels (L1, L2, L3), with L1 being the smallest and fastest, typically dedicated to data and instructions separately, while L3 is larger and slower but still quicker than RAM. The largest cache, usually called the last-level cache (LLC), if missed, requires fetching data from RAM. In the case of a multicore processor (as used in our experiments), each core typically has its own L1 cache, whereas higher-level caches might be shared among cores.

When the CPU needs data, it initially checks the fastest, smallest cache (L1). If the data is not there (known as a cache miss), it proceeds to the next cache level, continuing until it reaches the RAM. If the data is not found in the RAM, it must be retrieved from the disk, which is substantially slower. This hierarchical setup efficiently manages trade-offs between access speed, cost, and storage capacity, ensuring that frequently used data is quickly accessible at higher (faster) levels while less frequently accessed data is stored on slower storage mediums.

\subsubsection{Effects on computation time}

Regarding execution time, we can observe that time increases as more jobs are run concurrently. There is an increase of around 20\% to 40\% in execution time when comparing the execution of a single job to 10 concurrent jobs. This difference becomes more pronounced as the number of clusters increases, underscoring that the computing environment can drastically impact the algorithms' running time (and thus the relative speedup).

\subsubsection{L1 Cache Misses}

In the heatmaps of the second row of Figure~\ref{fig:perf_heatmaps} we analyze the percentage of L1 cache misses, calculated as the number of load misses from the L1 data cache divided by the total loads. This metric appears unaffected by the concurrency of jobs, which is expected since the L1 cache is exclusive to each core. Cache misses in the data cache can primarily result from two issues: (1) the data not fitting into the cache, resulting in repeated transfers from other caches or memory, and (2) poor data locality. Poor data locality occurs when elements are accessed alternately instead of sequentially; these elements are likely not present in the cache, requiring additional time to retrieve them from other memory sources. Modern CPUs attempt to mitigate this issue by employing caching strategies that prefetch data into the cache before it is explicitly requested. This is based on predictions of sequential access. This prefetching typically lowers cache miss rates and boosts performance. 

Observations indicate that with a lower number of clusters, both accelerated \Kpp variants exhibit fewer cache misses. This outcome is expected as fewer points need to be visited and/or fewer distances need to be calculated, thus reducing the frequency of memory accesses. Initially, when there are few clusters, the data access pattern retains some sequentiality. However, as more clusters are formed, points are less visited sequentially by the accelerated \Kpp variants. In other words, while the standard \Kpp variant continues to visit points sequentially (as it calculates distances from all points), the accelerated \Kpp variants only visit points within clusters that bypass the filters. Consequently, as the number of clusters increases, data access becomes less ordered, leading to increased cache misses in the accelerated algorithm variants. This trend is particularly pronounced in the norm-filtered version, which ultimately achieves a $25\%$ higher miss rate than the accelerated \Kpp TIE-only variant. This is probably because cluster points are further divided into two partitions, resulting in a significant increase in cache misses, likely impacting the norm-filtered version's computation time performance.

\subsubsection{Last-level cache misses}

In the third row, which analyzes last-level cache (LLC) misses, we observe that running multiple processes simultaneously affects the LLC miss rate. This outcome is expected, because the LLC is shared among all cores within the same CPU; thus, the actions of one core can impact the memory usage of another. As a result, the number of LLC misses increases with the number of concurrent jobs. This increase is more pronounced with a higher number of clusters.

Focusing on results from running a single job, where there is no external influence from other cores, we see a similar pattern as observed with L1 cache misses, especially when the number of clusters is low (e.g., $k=32$). The standard \Kpp variant initially exhibits more misses than the accelerated algorithm variants. However, the miss rate in the standard version decreases rapidly, reaching a low of about 4\%. In contrast, the accelerated \Kpp variants result in a less pronounced decrease. This holds in particular for the full accelerated \Kpp variant, which only shows a total reduction of 4\%.

The decrease in miss rates is expected as most of the data is likely loaded in the initial iterations, and as less of that data is used in subsequent iterations, it becomes more probable that the data can be stored in the cache or is already present. This scenario is optimal when accessing data sequentially, as in the standard \Kpp variant, because L2 and L3 caches, being larger than L1, can store more data and are more likely to have subsequent data pre-loaded. Conversely, non-sequential access is less efficient because it often necessitates accessing RAM, which is slower. In such cases, prefetching becomes more challenging. Particularly for the full accelerated \Kpp variant, which has a miss rate double as high as the one of the TIE-only variant, the issue of main memory access significantly impacts performance. While not shown here due to space constraints, we also observed a higher number of minor page faults in scenarios with a high number of clusters for both the standard and the full accelerated \Kpp variants. Minor page faults occur when a program accesses a page that is not loaded into RAM but is still available within the system’s virtual memory. Although this does not involve disk I/O operations, it introduces additional overhead.

As the number of concurrent jobs increases, we observe that the miss rate increases across all three algorithms. Although the full accelerated \Kpp consistently shows the highest miss rate, the differences to the TIE-only variant become less significant in proportion. In summary, when many processes are running on the same CPU, the LLC miss rate is expected to increase dramatically, thus affecting the running time of the algorithms.

\subsubsection{Number of instructions per cycle}

The number of instructions per cycle (IPC) measures the number of instructions executed per CPU cycle, with higher values indicating a more efficient algorithm for memory access. In this analysis, the standard \Kpp consistently achieves a higher IPC than the accelerated \Kpp variants. Hereby, the difference grows as the number of clusters increases, with the standard algorithm variants' IPC more than doubling that of the accelerated algorithm variants at $k=4096$. Additionally, the IPC decreases in all cases as the number of concurrent jobs increases.

One significant factor influencing IPC is memory latency. When the processor awaits data retrieval, it generally cannot execute further instructions until the data is fetched. Therefore, high latency, often due to cache misses, can significantly reduce IPC. As observed, the LLC experiences more misses as the number of concurrent jobs increases, adversely affecting the IPC.

However, the IPC measure of the standard \Kpp variant typically increases with the number of clusters. This increase can be attributed to both the sequential access of data and the reduction in calculations since points chosen, as centers do not need to be revisited, resulting in lower memory usage. At $k=4096$, the number of clusters accounts for approximately $1\%$ of the total number of points in this instance, which likely impacts data fetching during the final phases.

Conversely, IPC in both accelerated versions decreases as the number of clusters increases. This decrease is logical because although many calculations are saved (thus reducing memory access), the non-sequential nature of memory access in these versions introduces greater latency due to cache misses, causing the CPU to wait longer for data retrieval. The situation worsens with an increase in the number of concurrently running jobs on the same CPU. Hereby, the full accelerated \Kpp variant exhibits a lower IPC measure than the TIE-only variant across most scenarios, but these differences diminish with a higher number of concurrent jobs and clusters.

After evaluating the memory performance of the algorithms, it becomes apparent that the full accelerated \Kpp, despite reducing the number of calculations, achieves less speedup than expected, likely due to its poor data locality. A similar issue affects both accelerated \Kpp variants when compared to the standard one. Consequently, developing a more cache-friendly algorithm could potentially enhance the speedup of the accelerated algorithm variants even further.

\section{Conclusions and Future Work}

In this paper, we proposed an accelerated version of the exact \Kpp algorithm, leveraging geometric information, specifically the Triangle Inequality and additional norm filters, along with a two-step sampling procedure. Our experiments showed that the accelerated algorithm variants outperform the standard \Kpp in terms of the number of visited points and distance calculations, achieving greater speedup as the number of clusters increases. The Triangle Inequality-based acceleration is particularly effective for low-dimensional data, while the norm-based filter enhances performance in high-dimensional instances with a significant norm variance among points. Additional experiments demonstrated the behavior of our algorithms when executed concurrently across multiple jobs and examined how memory performance impacts the speedup measured in practice.

For future work, we plan to enhance the method by avoiding the calculation of all center-center distances at each iteration and improving the norm methods by using reference points other than the origin. Additionally, we aim to improve the algorithm by accessing data in a more ordered manner to enhance data locality and reduce cache memory failures, thereby aligning speedup more closely with the number of saved calculations. Some of these ideas are briefly introduced in appendices~A and ~B. 

\section*{Acknowledgements}
Guillem Rodríguez Corominas acknowledges support from the Department of Research and Universities of the Government of Catalonia by means of an ESF-founded pre-doctoral grant of the Catalan Agency for Management of University and Research Grants (AGAUR), under ref. number~
2022 FI\_B 00903. Maria~J.~Blesa acknowledges support from AEI under grant PID-2020-112581GB-C21 (MOTION). Christian Blum was supported by two grants funded by MCIN/AEI/10.13039/501100011033: TED2021-129319B-I00 and PID2022-136787NB-I00. \\



\appendix

\section{Avoiding center-center distance computations}
\label{sec:appendix:avoidingDistances}

As explained, both accelerated algorithm versions require calculating the distances between the newly selected center and the existing ones.
Although this step does introduce additional computational overhead, it is crucial for reducing the far greater number of distance calculations that would otherwise be necessary between each point and the new center, especially given that, in practice, the number of points usually greatly exceeds the number of clusters.

However, particularly when the number of clusters is large, calculating the distance between the new center and previous ones can become unnecessary. As new clusters are formed, they tend to decrease in size. Furthermore, when two clusters are significantly separated, it becomes apparent that points within one cluster are unlikely to be closer to a point from another cluster, even if that point becomes the newly selected center, than they are to their existing center. In such scenarios, we can, again, make use if the TIE in order to avoid calculating the respective center distances. This can be especially beneficial in high-dimensional spaces where saving even a single distance calculation can significantly impact execution time.

More specifically, let $\mathcal{P}_1$ and $\mathcal{P}2$ be two clusters each with radii $r_1$ and $r_2$, respectively. Moreover, consider $\vv{c}_{new} = \vv{p}_i$ as the newly selected center from the first cluster, i.e., $\vv{p}_i \in \mathcal{P}_1$. Given the distance $d(\vv{c}_1, \vv{c}_2)$ between the centers of these two clusters we can derive the following using the TIE:
\begin{equation}
d(\vv{c}_{new}, \vv{c}_2) \geq d(\vv{c}_1, \vv{c}_2) - d(\vv{c}_{new}, \vv{c}_1),
\end{equation}
Note that $d(\vv{c}_{new}, \vv{c}_2)$ is equal to the upper bound of point $\vv{p}_i$ (the new center) and that both $d(\vv{c}_1, \vv{c}_2)$ and $d(\vv{c}_{new}, \vv{c}_1)$ have been calculated in previous iterations. Recall that if a point in $\mathcal{P}_2$ is to be closer to $\vv{c}_{new}$ than to $\vv{c}_2$, we require:
\begin{equation}
d(\vv{c}_{new}, \vv{c}_2) \leq 2 r_2.
\end{equation}
Combining these inequalities provides the condition
\begin{equation}
 d(\vv{c}_1, \vv{c}_2) - d(\vv{c}_{new}, \vv{c}_1) \geq 2 r_2,
\end{equation}
If this inequality holds, it means that all points in $\mathcal{P}_2$ are definitively closer to $\vv{c}_2$ than to $\vv{c}_{new}$, thus avoiding unnecessary distance calculations. If this condition is not met, then we have to proceed as normal and explicitly calculate the aforementioned center-center distance. This approach helps avoiding unnecessary computations, particularly in cases where cluster separations are large relative to their radii.

This principle can be extended to any point within $\mathcal{P}_1$. We know, by definition, that the radius $r_1$ is greater than $d(\vv{c}_{new}, \vv{c}_1)$ for every $\vv{c}_{new}$ in $\mathcal{P}_1$. Therefore, if the following condition is satisfied:
\begin{equation}
d(\vv{c}_1, \vv{c}_2) - r_1 \geq 2 r_2,
\end{equation}
it can be inferred that any point from $\mathcal{P}_2$ is closer to its current center than to any future center selected from $\mathcal{P}_1$. Note that, since $r_1$ and $r_2$ are non-increasing with each iteration, once the above inequality holds, it will continue to hold in all subsequent iterations. In the case of the accelerated version using the norm, as we are further dividing a cluster into two partitions, the larger radius between the two can be used. However, if only using the norm filter, pairwise distances between centers are not needed.

\section{Improving the norm filter}
\label{sec:appendix:improvingNorm}

As observed, the norm filter typically shows enhanced effectiveness when the norm variance is higher. However, these methods also introduce some overhead due to the norm calculations. In this section, we propose several improvements aimed at reducing this overhead while enhancing the effectiveness of the norm filter, especially in scenarios where it performs worse.

First, it is important to note that the norm of a point is equal to its Euclidean Distance (ED) to the origin, which serves as a default reference point. However, in practice, any point in the space can function as a reference point. Note that this is equivalent to shifting the data such that a new reference point is positioned at the origin, after which the norms are calculated as usual. In fact, shifting the data does not alter the relative distances among points, ensuring that the obtained results remain the same. When the norm variance is low, this might indicate that the points are unfavorably distributed in space relative to the origin. Thus, by strategically selecting a different reference point, essentially shifting the point of view, we can potentially enhance the effectiveness of the norm filter.

The process of selecting a new reference point, however, needs to be fast; otherwise, the additional time spent identifying it may negate the potential time savings during forthcoming iterations. For intance, a good reference point could be positioned along the principal axis, which is the line representing the direction of the principal component, i.e., where the data exhibits the greatest variance when projected onto it. However, calculating the principal component, especially in high-dimensional instances, can be computationally intensive and may not justify the extra efforts, making it potentially impractical in these cases.

\begin{table}[htb]
\caption{List of norm variance per instance (in \%) given different reference points.}
\label{tab:norms}
\centering
\resizebox{0.6\columnwidth}{!}{
\renewcommand{\arraystretch}{1.25}
\begin{tabular}{lrrrrr}

 Instance & Origin  &  Mean & Median & Positive & Mean Norm \\
 \hline
 MGT  & $\mathbf{50.00}$ & $28.26$ & $33.22$ & $18.40$ & $33.13$\\ 
\hline
CIF-C & $11.49$ & $11.94$ & $12.51$ & $\mathbf{25.37}$ & $18.48$ \\
 \hline
 CIF-T & $48.06$ & $32.37$ & $37.70$ & $22.42$ & $\mathbf{60.10}$ \\
 \hline
RQ & $2.60$ & $36.47$ & $52.17$ & $\mathbf{93.04}$ & $91.91$ \\ 
 \hline
 S-NS & $75.45$ & $18.75$ & $27.42$ & $\mathbf{75.52}$ & $15.81$\\ 
\hline
3DR & $22.63$ & $37.32$ & $46.88$ & $\mathbf{98.82}$ & $32.95$\\
 \hline
 RNA & $8.97$ & $27.07$ & $\mathbf{29.26}$ & $6.56$ & $21.31$\\
  \hline
 HPC & $5.40$ & $32.50$ & $63.27$ & $30.90$ & $\mathbf{70.52}$\\
 \hline
 HAR & $10.43$ & $19.26$ & $38.74$ & $15.71$ & $\mathbf{48.24}$\\ 
 \hline
 GS-MET & $56.38$ & $12.12$ & $12.52$ & $\mathbf{69.47}$ & $20.32$ \\
 \hline
 GS-CO & $\mathbf{85.12}$ & $23.80$ & $25.19$ & $84.56$ & $24.31$ \\
\hline
 YAH & $4.84$ & $9.07$ & $9.40$ & $0.40$ & $\mathbf{28.80}$\\ 
 \hline
\\
\hline
 GSAD  & $\mathbf{85.56}$ & $43.12$ & $61.49$ & $79.72$ & $58.30$\\
  \hline 
  PHY & $7.48$ & $4.42$ & $7.77$ & $9.32$ & $\mathbf{11.90}$\\ 
 \hline
 CRP & $52.92$ & $10.24$ & $11.65$ & $\mathbf{57.65}$ & $13.32$ \\ 
\hline
 C-10  & $23.61$ & $6.47$ & $6.86$ & $\mathbf{23.66}$ & $4.57$ \\  
 \hline
 C-100 & $28.08$ & $7.42$ & $8.01$ & $\mathbf{28.14}$ & $5.54$ \\ 
  \hline
 MNIST & $5.51$ & $1.82$ & $5.13$ & $\mathbf{5.55}$ & $4.14$ \\ 
 \hline
PTN & $\mathbf{85.12}$ & $55.87$ & $63.70$ & $76.66$ & $56.80$\\ 
\hline
 YP & $\mathbf{61.49}$ & $31.47$ & $38.88$ & $4.11$ & $26.29$\\
 \hline
 SUSY & $\mathbf{20.96}$ & $11.06$ & $13.08$ & $9.30$ & $15.15$ \\
 \hline
 
\end{tabular}
}
\end{table}

Here, we provide some examples of various points that could be selected as reference points. Table~\ref{tab:norms} displays the norm variance for several reference points, which are: 
\begin{itemize}
    \item Origin: The origin (note that using the origin as the reference point matches the standard calculation of the norm).
    \item Mean: The mean position of the points.
    \item Median: The median of the points.
    \item Positive: Selecting a reference point at the lower bounds of the bounding box that encapsulates all points, thereby ensuring that every coordinate value is non-negative. This is equal to shifting all data points such that the minimum value in each dimension is zero, thus relocating them to the positive quadrant.
    \item Mean Norm: The point whose norm is closest to the mean norm of the original data.
\end{itemize}
Additionally, the best value for each instance is highlighted in bold. Instances are ordered by size and divided into low and high-dimensional groups for a clearer comparison.

One notable observation is that selecting a different reference point exhibits significantly more benefits in low-dimensional instances compared to high-dimensional ones. For example, in high-dimensional instances, the original norm still achieves the highest norm variance in almost half of the cases, with the largest observed improvement from changing the reference point being only about $5\%$. In general, using the mean or median as the reference point appears to be less effective, as other points consistently provide better norm variances, with only one exception. In all cases, choosing the median as the reference point outperforms the mean, likely due to the latter being less sensible to outliers.

In lower-dimensional instances, we observe significantly greater benefits from changing the reference point, with improvements reaching up to around $90\%$ in some cases. Notably, the largest improvements are typically seen in instances with a lower original norm variance. For example, in instances where the variance was initially greater than 50\%, changing the reference point usually results in negligible gains or only minor improvements. For instance, instance S-NS achieved a mere $0.07\%$ gain, which is practically negligible given the additional computational effort that would be required for shifting the data. Conversely, other instances like CIF-T and GS-MET obtain gains of approximately 12\% and 13\%, respectively.

However, in cases where the original norm variance was lower, changing the reference point led to much more substantial increases. It appears that shifting data to the positive quadrant generally increases norm variance in these scenarios. Moreover, using the point whose norm is closest to the mean norm as the reference point can be advantageous in situations where the first strategy is less effective.

Thus, changing the reference point proves beneficial, especially in lower-dimensional instances with initially low norm variances. These instances are often those where the accelerated version utilizing both filters underperforms, and changing the reference point can significantly enhance its effectiveness. Ultimately, it is advisable for users, who have a deeper understanding of their data distribution, to determine the most suitable reference point, as its effectiveness heavily relies on it. We plan to further investigate this aspect of reference points, with the goal of identifying quick and optimal choices for reference point that enhances overall performance.

Additionally, the calculation of distances can be optimized. Let $x$ and $y$ be two points. Recall that the Squared Euclidean Distance (SED) between these two points is defined as follows:
\begin{equation}
    \SED(\vv{x},\vv{y}) = \norm{\vv{x} - \vv{y}}^{2} = \sum_{j=1}^{d} (\vv{x}_j - \vv{y}_{j})^2
\end{equation}
which can also be expressed using the norms as
\begin{equation}
    \SED(\vv{x},\vv{y}) = \norm{\vv{x}}^2 + \norm{\vv{y}}^2 - 2x \cdot y
\end{equation}
where $\cdot$ denotes the dot product. Note that the squared norms can be pre-computed at the beginning of the process (along with the computation of the norms). Furthermore, in out context, the new center is always involved in distance calculations at a given iteration (both center-center and point-center distances). Thus, it is also feasible to pre-compute the component $2x$ for each new center at the beginning of each iteration and reuse it throughout. Consequently, the remaining computational requirement for each distance calculation is primarily the dot product. This approach, then, reduces the number of operations required for each distance calculation.





\end{document}